\documentclass{ecai}
\usepackage{graphicx}
\usepackage{latexsym}
\usepackage{times}
\usepackage{soul}
\usepackage{url}
\usepackage{amsmath}
\usepackage[hidelinks]{hyperref}
\usepackage[utf8]{inputenc}
\usepackage[small]{caption}
\usepackage{graphicx}
\usepackage{stfloats}
\usepackage{amsthm}
\usepackage{booktabs}
\usepackage{algorithm}
\usepackage{algorithmic}
\usepackage{amssymb}
\usepackage{multirow}
\usepackage{bbding}
\usepackage{makecell}
\usepackage{booktabs}
\usepackage{tabu}  
\usepackage{bm}
\usepackage{graphicx}
\usepackage{subfigure}
\usepackage{color} 
\usepackage{enumitem}
\usepackage{colortbl}
\usepackage{arydshln}
\usepackage[most]{tcolorbox}  
\usepackage{amsthm} 	
\usepackage[switch]{lineno}
\usepackage{caption}
\usepackage{tcolorbox}
\usepackage[marginal]{footmisc}
\definecolor{darkgrey}{RGB}{120,120,120}
\definecolor{mygrey}{RGB}{200,200,200}
\let\oldalign\align
\let\oldendalign\endalign
\renewenvironment{align}
  {\linenomathNonumbers\oldalign}
  {\oldendalign\endlinenomath}

\let\oldequation\equation
\let\oldendequation\endequation
\renewenvironment{equation}
  {\linenomathNonumbers\oldequation}
  {\oldendequation\endlinenomath}

\newtheorem{definition}{Definition}

\def \hA {{\hat{A}}}

\let\emph\textit

\begin{document}

\begin{frontmatter}

\title{Adversarial Erasing with Pruned Elements: Towards Better Graph Lottery Tickets}

\author[A;\textdagger]{\fnms{Yuwen}~\snm{Wang}}
\author[A;\textdagger]{\fnms{Shunyu}~\snm{Liu}}
\author[A;\textdagger]{\fnms{Kaixuan}~\snm{Chen}}
\author[A]{\fnms{Tongtian}~\snm{Zhu}}
\author[B]{\fnms{Ji}~\snm{Qiao}}
\author[B]{\fnms{Mengjie}~\snm{Shi}}
\author[A;*]{\fnms{Yuanyu}~\snm{Wan}} 
\author[A]{\fnms{Mingli}~\snm{Song}}
\address[A]{Zhejiang University, Hangzhou, China}
\address[B]{China Electric Power Research Institute, Beijing, China}

\begin{abstract}
Graph Lottery Ticket~(GLT), a combination of core subgraph and sparse subnetwork, has been proposed to mitigate the computational cost of deep Graph Neural Networks~(GNNs) on large input graphs while preserving original performance.
However, the winning GLTs in exisiting studies are obtained by applying iterative magnitude-based pruning~(IMP) without re-evaluating and re-considering the pruned information, which disregards the dynamic changes in the significance of edges/weights during graph/model structure pruning, and thus limits the appeal of the winning tickets. 
In this paper, we formulate a conjecture, i.e., existing overlooked valuable information in the pruned graph connections and model parameters which can be re-grouped into GLT to enhance the final performance.
Specifically, we propose an~\emph{adversarial complementary erasing}~(ACE) framework to explore the valuable information from the pruned components, thereby developing a more powerful GLT, referred to as the \textbf{ACE-GLT}. The main idea is to mine valuable information from pruned edges/weights after each round of IMP, and employ the ACE technique to refine the GLT processing. Finally, experimental results demonstrate that our ACE-GLT outperforms existing methods for searching GLT in diverse tasks. Our code is available at \url{https://github.com/Wangyuwen0627/ACE-GLT}.
\end{abstract}
\end{frontmatter}

\renewcommand{\thefootnote}{\fnsymbol{footnote}}
\footnotetext[2]{Equal Contribution. Email: \{yuwenwang, liushunyu, chenkx\}@zju.edu.cn.}
\footnotetext[1]{Corresponding Author. Email: wanyy@zju.edu.cn.}
\renewcommand{\thefootnote}{\arabic{footnote}}

\section{Introduction}
\label{section1}
{Graph Neural Networks~(GNNs)~\cite{kipf2016semi,velivckovic2017graph,xu2018powerful,jing2021amalgamating} have emerged as the leading architectures to analyze non-Euclidean samples where information is present in the form of graph structure, and have shown encouraging performance in various downstream tasks such as node classification~\cite{velivckovic2017graph}, link prediction~\cite{zheng2020robust}, and graph classification~\cite{chen2022DKEPool}. However, the high computational bottleneck, which is predominantly caused by the dense connectivity in GNNs and the larger-scale graph samples used as input, decreases the capability of the feature aggregation during the GNNs' training. Moreover, these inherent shortcomings limit the application of GNNs in large-scale tasks, especially in resource-constrained cases.}

{Recently, there has been a surge of works focusing on mitigating this inefficiency~\cite{rong2019dropedge,tailor2020degree,chen2021unified,you2022early} by exploring the masked adjacency matrix and sparse model parameters, respectively. In particular, the Graph Lottery Ticket~(GLT)~\cite{chen2021unified} is a combination of core subgraph and subnetwork achieved by pruning the graph edges and model parameters simultaneously, maintaining a consistent level of accuracy compared to the non-pruned one.
However, these existing methods employ the irreversible pruning scheme, which means that edges and parameters are removed at the start of the iterative magnitude-based pruning (IMP) procedure will not be taken into account in subsequent selections.
Furthermore, this scheme fails to consider a critical factor, i.e., the importance of the edges and model weights is dynamically changing during the IMP procedure.

\begin{figure}[!t]
\includegraphics[width=1.0\linewidth]{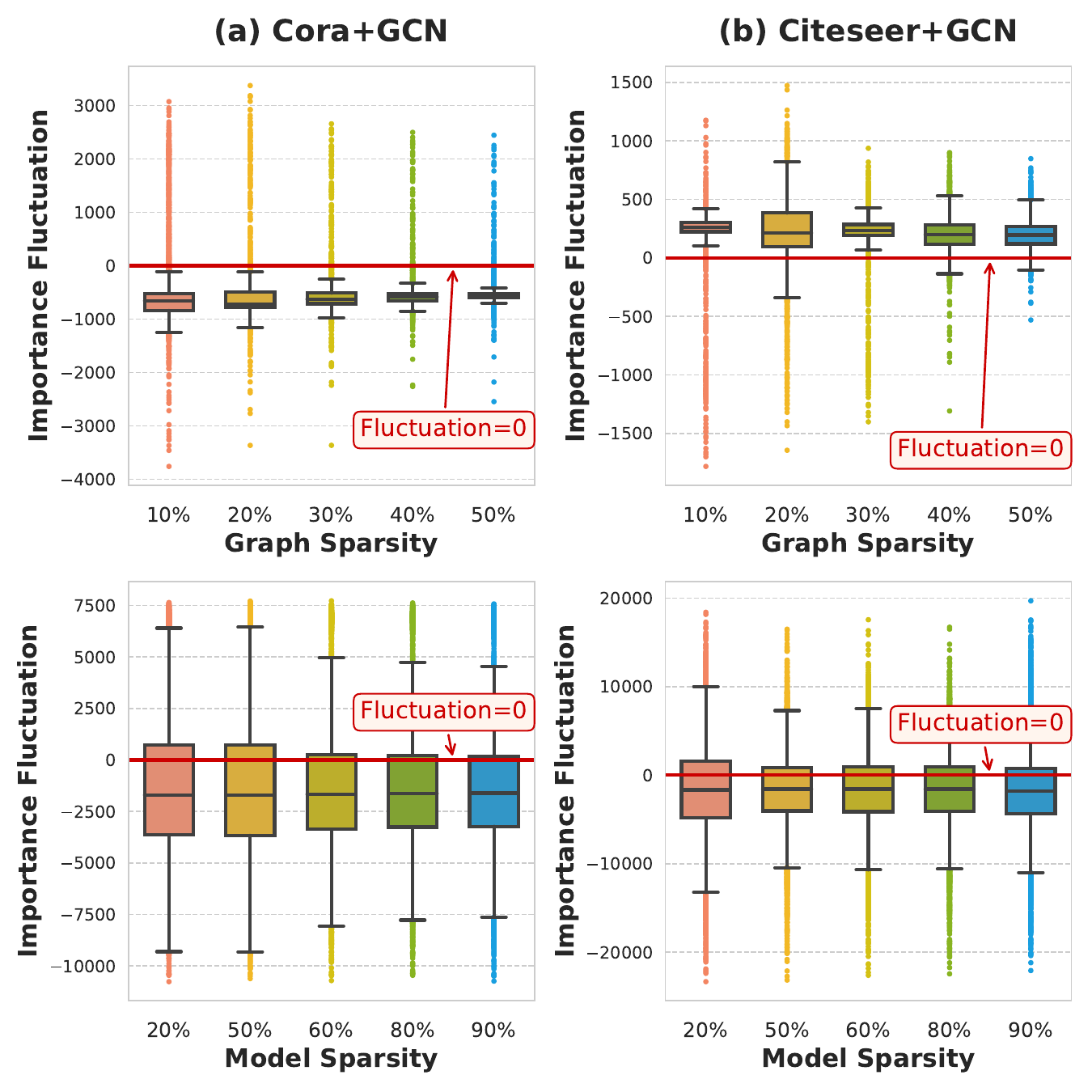}
\captionsetup{justification=centering}
\caption{The illustration of importance fluctuation regarding the graph edges~(upper row) and model parameters~(lower row) during IMP process. The farther the importance fluctuation is from zero, the more severe the dynamic change.
}
\label{fig:fig1}
\end{figure}
To illustrate the importance fluctuation\footnote{The importance fluctuation measures the changes in the importance ranking of the winning GLT's elements compared to that at different IMP stages.} of graph edges and model parameters during the IMP process, we present the Figure~\ref{fig:fig1}, which collects the importance fluctuation of edges/weights elements on the Cora and Citeseer benchmarks, with GCN serving as the baseline algorithm. The horizontal coordinates correspond to varying levels of sparsity, which indicate the corresponding stages in the iterative pruning process, and the vertical coordinates represent importance fluctuation of the resulting elements from IMP process. 
As shown in Figure~\ref{fig:fig1}, the importance fluctuation, which is farther from zero, reveals the noticeable dynamic change of the retained edges or weights resulting from the IMP pruning.
Therefore, we can infer that such irreversible IMP scheme will lead to two following issues: (1) The irreversible IMP scheme may cause valuable information to be permanently deleted due to the overlooked importance fluctuation of edge and weight elements. (2) As a result, the compression of the graph structure and model  parameters is inevitably inadequate, leading to a significant drop in the performance.
}

\begin{figure*}[!ht]
\includegraphics[width=1.0\linewidth]{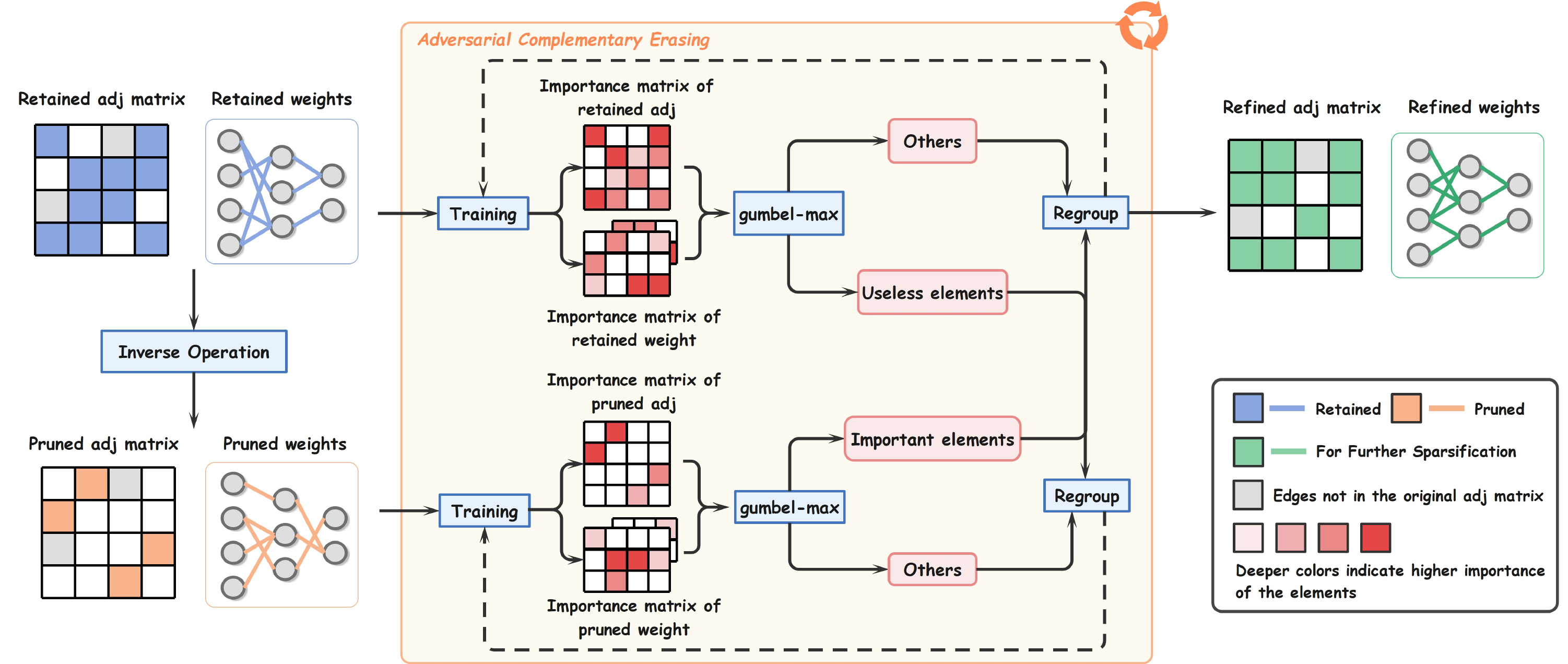}
\captionsetup{justification=centering}
\caption{An illustration of the proposed ACE-GLT framework. First, we use magnitude-based pruning to obtain the retained substructure and take the inverse of it to represent the pruned substructure. Then, we iteratively extract valid information from the pruned part to refine the retained part. Finally, the refined substructure is used as a target for further sparsification.}
\label{fig:fig5}
\end{figure*}

{To mitigate the adverse effects of edges/weights importance fluctuation during the IMP process, we suggest to take all previously pruned elements into consideration and re-evaluate the current importance before further sparsification. The central concept is to mine the valuable information of the pruned elements after each round of IMP, and employ the achieved knowledge to refine the GLT processing. Thus, the challenge is:
\begin{tcolorbox}[notitle, sharp corners, colframe=darkgrey, colback=white, 
       boxrule=1pt, boxsep=0.5pt, enhanced, 
       shadow={3pt}{-3pt}{0pt}{opacity=1,mygrey},
       title={Challenge},]
\emph{{How to design an effective framework to purify valuable information from the pruned elements to the retained GLT?}}
\end{tcolorbox}

}

In this paper, we present the \emph{Adversarial Complementary Erasing} (ACE) optimization framework to recover the missed "lucky numbers", i.e., the valuable elements that are mistakenly pruned by irreversible IMP scheme.
Then, we iteratively apply the ACE framework after each round of pruning, refining the GLT search process and producing a higher-performing GLT, and thus name our proposed method as \textbf{ACE-GLT}.
Specifically, as shown in Figure~\ref{fig:fig5}, we first apply magnitude-based pruning to obtain the retained substructures and take their inversion to represent the pruned ones. Then, to refine the retained substructures, we implement adversarial complementary erasing by iteratively locating and exchanging the most disruptive parts of the retained substructure and the most discriminative parts of the pruned substructure through gumbel-max sampling. In this way, the surplus value will be maximally squeezed from pruned part to retained parts. Finally, we adopt the retained substructure after refinement as the target for further sparsification.

The contributions of this paper are summarized below:
\begin{itemize}
    \item We explore the problem of significant importance fluctuation of elements resulting from the dynamic pruning of the graph/model structure for the first time. Then we propose to alleviate the problem by re-considering pruned elements and mining valuable information from pruned parts.
    \item We propose an adversarial complementary erasing method for graph lottery ticket (ACE-GLT) framework to find the approximate-optimal GLT by compressing the residual value of pruned parts in the process of IMP.
    \item Extensive experiments are conducted on GNN benchmarks to examine our method. The results show that our method consistently outperforms those GLT-based approaches over these benchmarks. These results also demonstrate the potential of ACE-GLT to uncover additional valuable information in pruned parts.
\end{itemize}

\section{Related Work}
The computational bottleneck of GNNs has received much research interest. Most existing studies are concerned about the high computational cost caused by large input graphs and thus investigate techniques of graph simplification, including graph sparsification and graph coarsening, to improve the efficiency of GNNs. Specifically, graph sparsification aims to extract a small subgraph from the large input graph while keeping the performance for learning tasks. For example, DropEdge~\cite{rong2019dropedge} randomly removes edges from the original graph during the training process, enhancing the randomness and diversity of the input data while reducing the computational cost of the message-passing process. SGCN~\cite{li2020sgcn} attempts to formulate sparsification as an optimization problem and effectively solve it via the alternating direction method of multipliers. PDTNet~\cite{luo2021learning} prunes task-irrelevant edges by penalizing the number of edges in the sparse graph with parameterized deep neural networks and imposes a low-rank constraint on the sparse graph with nuclear norm regularization to achieve better generalization. In contrast to graph sparsification techniques, graph coarsening aims to group original nodes into super-nodes and re-define their connections based on similar constraints~\cite{loukas2018spectrally,loukas2019graph,deng2019graphzoom}. DiffPool~\cite{ying2018hierarchical} proposes a hierarchical graph clustering approach, where each layer maps the nodes of this layer into several clusters and takes the clustering results as input nodes for the next layer. GCOND~\cite{jin2021graph} condenses the large graph into a small and highly-informative by imitating the GNN training trajectory on the original graph through the optimization of a gradient matching loss and designs a strategy to condense node features and structural information simultaneously. In addition, SAGPool~\cite{lee2019self} uses graph convolution to compute a self-attention score with node features and graph topology to identify the nodes to be removed.

However, with the emergence of deeper GNN architectures~\cite{li2020deepergcn,chen2020simple,min2020scattering}, only applying graph simplification is insufficient to tackle the computational bottleneck of GNNs because the enormous number of model parameters also brings unacceptable costs.
Recent study\cite{frankle2018lottery} has suggested the lottery ticket hypothesis~(LTH) to tackle the problem of redundancy in model parameters, specifically in image classification tasks.
The LTH claims the existence of a winning ticket (i.e., a
properly pruned sub-network together with original weight initialization) that can achieve competitive performance to the original dense network. The following work~\cite{frankle2019stabilizing} attempts to obtain a robust pruned subnetwork by resetting the weights to those obtained after a few training rounds, instead of using the model initialization. Furthermore, the work~\cite{renda2020comparing} extends the training of the subnetwork from initialization to the early stage of pretraining, which improves the accuracy of the subnetwork in more challenging tasks.
Apart from the image classification task, the LTH is also expanded into many other research areas. The work~\cite{girish2021lottery} proposes a guidance for the identification of task-specific winning tickets for
object detection, instance segmentation, and keypoint estimation. Data-LTH-ViTs~\cite{Shen2022lottery} extends LTH to ViT by finding a subset of input image patches as data-level winning tickets. Besides, TAMT~\cite{liu2022learning} learns to find winning tickets in BERT Transfer through task agnostic mask training.

The work ~\cite{chen2021unified} extends LTH to the non-Euclidean domain for the first time and introduces a new concept called graph lottery ticket~(GLT). GLT is defined as a combination of sparse subgraph and subnetwork extracted from original graphs and GNNs, which can achieve comparable or even exceeding accuracy compared with non-pruned ones. It jointly prunes model parameters and graph connections to accelerate GNNs inference effectively. Based on this work, CGP~\cite{liu2022comprehensive} further incorporates the sparsification of node features into GLT. Moreover, ICPG~\cite{sui2022inductive} points out that UGS is incapable of inductive learning. It proposes a graph encoder to learn connection masks and successfully identifies GLTs on unseen graphs. In addition, UGT~\cite{huang2022you} focus on the compression of model parameters and finds a well-performing subnetwork without any training of the model.
A more recent work~\cite{wang2023searching} proposes DGLT to prune the GNN parameters by incremental regularization~\cite{wang2021neural} and hierarchical sparsification on input graph, which enables the search of GLT from a dual perspective. In addition to the general GLT, GEBT~\cite{you2022early} proves that the early-bird ticket, a winning ticket that can be extracted in the early stages of training~\cite{you2019drawing}, also exists in GNNs. Despite extensive research on GLT, previous studies employ the irreversible pruning scheme and disregard the impact of dynamic changes in structure on importance assessment of elements. This oversight may cause the omission of critical subgraphs and subnetworks during the process of progressive sparsification. In this paper, we introduce an adversarial complementary erasing framework to recover omitted information from the pruned part and find a GLT with better performance. 

\section{Method}
In this section, we first present the essential preliminaries of graph neural networks~(GNNs) and graph lottery tickets~(GLTs). Then, we elaborate on our method of adversarial complementary erasing~(ACE) for GLT, which we refer to as ACE-GLT.
\subsection{Preliminaries}
Let $\mathcal{G}=\{{A},{X}\}$ denote an undirected graph with $n$ nodes, where ${A}\in \{0,1\}^{n \times n}$ is the adjacency matrix and ${X} \in \mathbb{R}^{n \times d}$ is the matrix consisting of the $d$-dimensional feature vector of each node. Then, let $f(\cdot; {W})$ denote a GNN model, where ${W}$ represents parameters of the model. For example, a two-layer GCN model can be formulated as follows:
\begin{equation}
\label{eq1}
\begin{aligned}
Z = f(\{{A}, {X}\}; {W}) =\operatorname{Softmax}(\hA\sigma (\hA {X} {W}^{(0)}) {W}^{(1)}),
\end{aligned}
\end{equation}
where $Z$ is the output of the model, $\hA={\hat{D}}^{-\frac{1}{2}}({A}+{I}_n) {\hat{D}}^{-\frac{1}{2}}$ is the normalized adjacency matrix, $\tilde{{A}}={A}+{I}_n$ is the adjacency matrix with added self-loop, $\hat{{D}}$ is the degree matrix of $\tilde{{A}}$, $\sigma(\cdot)$ is an activation function, and ${W}^{(k)}$ represents the trainable weight matrix of the $k$-th layer in model.
Moreover, for an input graph $\mathcal{G}$ and a GNN model $f(\cdot;{W})$, 
let $\mathcal{G}_{sub}=\{{A} \odot {M}_A,{X}\}$ denote a subgraph of $\mathcal{G}$ and $f_{sub}(\cdot; {W} \odot {M}_W)$ denote a subnetwork of $f(\cdot;{W})$, where ${M}_{A}$ and ${M}_{W}$ represent the mask matrices of adjacency matrix and model weights, $\odot$ denotes the element-wise product.
With $\mathcal{G}_{sub}$ and $f_{sub}$, the graph lottery ticket~(GLT) is defined as follows.

\begin{definition}[Graph Lottery Ticket]\label{def:glt}
  \emph{Given an input graph $\mathcal{G}$, a GNN model $f(\cdot;{W})$ and initialization of model parameters ${W_{init}}$, a graph lottery ticket is a combination of sparse graph $\mathcal{G}_{sub}$ and sparse model $f_{sub}$, which can match the test accuracy of the non-pruned case when trained in isolation for at most the same number of iterations with dense model.}
\end{definition}

\subsection{ACE-GLT}
In the following, we first explain our motivations, and then present our ACE framework as well as its application in discovering GLT.\\

\noindent\textbf{Motivations.}
As previously discussed, GLT can be found by iteratively invoking a magnitude-based pruning algorithm~\cite{chen2021unified}. For better description, detailed procedures of the algorithm are outlined in Algorithm~\ref{alg:alg1}. First, for a graph $\mathcal{G}=\{{A},{X}\}$ and a GNN model $f(\cdot;{W})$ initialized by ${W_{init}}$, the masks of adjacency matrix and model weights are initialized as follows:
\begin{align}
{M_A^0}={A}, {M_W^0}=\mathbf{1} \in \mathbb{R}^{\left\|{W_{init}}\right\|_0}.
\end{align}
Then as shown in Line~1--4 of Algorithm~\ref{alg:alg1}, these masks are softened to be differentiable and updated along with the model weights based on the loss $\mathcal{L}_\text{retained}$:
\begin{equation}
\label{eq2}
\begin{aligned}
\mathcal{L}_\text{retained}:=  \mathcal{L_\text{ce}}&(f_{sub}(\{{A}\odot{M_A}, {X}\};{W} \odot {M_W}),y)\\
&+\lambda_1\left\|{M_A}\right\|_1+\lambda_2\left\|{M_W}\right\|_1,
\end{aligned}
\end{equation}
where $\mathcal{L}_\text{ce}$ represents the cross-entropy loss, $y$ is the true label, and $\lambda_1$, $\lambda_2$ control $\ell_1$ regularization of masks. Then, the masks converge to ${M_A^n}$ and ${M_W^n}$ through $n$ rounds of iterative training. Finally, as shown in Line~5--6 in Algorithm~\ref{alg:alg1}, the elements with lower magnitude in ${M_A^n}$ and ${M_W^n}$ are set to zero, w.r.t. pruning ratios $p_A$ and $p_W$, and the rest of the elements are set to one. The modified masks are ultimately outputted as ${M_A}$ and $ {M_W}$.

After each run of Algorithm~\ref{alg:alg1}, the output masks will be used to sparsify ${A}$ and ${W}$, then the retained part can be denoted as: 
\begin{equation}
\begin{aligned}
{Z}_\text{retained}=f_{sub}(\{{A} \odot {M_A},{X}\};{W_{init}} \odot {M_W}). 
\end{aligned}
\end{equation}

The existing studies obtain GLT by iteratively invoking Algorithm~\ref{alg:alg1} to prune the retained parts until the specified sparsity is achieved. However, in these studies, pruned elements of the current iteration will not be considered in subsequent processes. We have verified that the elements pruned early may be crucial for post-pruning substructure in Section~\ref{section1}. Therefore, the irreversible pruning mechanism adopted by previous methods undoubtedly results in the loss of some valuable information, leading to a reduction in the performance of the found GLT. To mitigate performance degradation, we develop an adversarial complementary erasing (ACE) framework, which explicitly exploits the pruned part to refine the masks.\\

\begin{algorithm}[t]
\caption{Magnitude-based Pruning}
\label{alg:alg1}
\begin{algorithmic}[1]
\REQUIRE Graph $\mathcal{G}=\{{A},{X}\}$, GNN's initialization ${W_{init}}$, GNN $f(\cdot;{W_{init}})$, pruning ratios of each iteration ${p_A}$ for graphs and ${p_W}$ for GNNs, initial masks ${M_A^0}={A}$, ${M_W^0}=\boldsymbol{1} \in \mathbb{R}^{\left\|{W_{init}}\right\|_0}$.
\ENSURE Sparsified masks: ${M_A}$ and ${M_W}$ 
\FOR{iteration $i = 0, 1, 2, ..., n-1$}
\STATE Forward $f_{sub}(\{{A}\odot{M_A^i},{X}\};{W}_i\odot{M_W^i})$ to compute the loss $\mathcal{L}_{s}$ in Equation~\eqref{eq2}.
\STATE Update ${W}_{i+1}$, ${M_A^{i+1}}$ and ${M_W^{i+1}}$ according to $\mathcal{L}_{s}$.
\ENDFOR
\STATE Set ${p}_A$ of the lowest-magnitude values in ${M_A^n}$ to 0 and others to 1, then obtain ${M_A}$.
\STATE Set ${p}_W$ of the lowest-magnitude values in ${M_W^n}$ to 0 and others to 1, then obtain ${M_W}$.
\end{algorithmic}
\end{algorithm}

\noindent\textbf{Adversarial Complementary Erasing.}
Before introducing the ACE framework, we first denote the pruned part after each invocation of Algorithm~\ref{alg:alg1} as ${Z}_\text{pruned}$, which can be computed as follows:
\begin{align}
&{M_A^r} = {A}\oplus{M_A},{M_W^r} = \lnot {{M_W}},\\
&{Z}_\text{pruned}=f_{sub}(\{{A} \odot {M_A^r},{X}\}; {W_{init}} \odot {M_W^r}),
\end{align}
where
$\oplus$ denotes the element-wise XOR (exclusive or) operator, and $\lnot$ denotes the element-wise logical not operator.

Our goal is to remove the impurities from the retained parts ${Z_\text{retained}}$ and to ``pan for gold'' from the pruned part ${Z_\text{pruned}}$ at the same time. The proposed method achieves this goal by iteratively performing an adversarial complementary erasing process, which replaces some elements of the retained part ${Z_\text{retained}}$ with elements of the pruned part ${Z_\text{pruned}}$. Specifically, in each round $i$, ${Z}_\text{retained}$ and ${Z}_\text{pruned}$ are first trained separately. During the training process, ${M_A},{M_W}$ are optimized according to Equation~\eqref{eq2}, and ${M_A^r},{M_W^r}$ are optimized according to $\mathcal{L}_{\text{pruned}}$:

\begin{equation}
\label{eq3}
\begin{aligned}
\mathcal{L}_{\text{pruned}}:=  \mathcal{L_\text{ce}}&(f_{sub}(\{{A}\odot{M_A^r}, {X}\};{W} \odot {M_W^r}), y)\\
&+\lambda_1\left\|{M_A^r}\right\|_1+\lambda_2\left\|{M_W^r}\right\|_1,
\end{aligned}
\end{equation}
where the first term represents the cross-entropy loss, and $\lambda_1$, $\lambda_2$ are hyparameters to control $\ell_1$ sparsity regularization of mask matrices. Notably, the elements in mask matrices are soften to continuous values in $[0, 1]$ while training.

\begin{algorithm}[!t]
\caption{Adversarial Complementary Erasing} 
\label{alg:alg2}
\begin{algorithmic}[1]
\REQUIRE GNN's initialization ${W_{init}}$, Graph $\mathcal{G}=\{{A},{X}\}$, GNN $f(\cdot, {W_{init}})$, Adversary rounds $T$, Upper limit of sampling $\mathcal{K}$.
\ENSURE Refined masks: ${M_A}^{\prime}$ and ${M_W}^{\prime}$
\STATE Sparsify GNN ${Z}$ with graph $\mathcal{G}$ and obtain ${M_A}$, ${M_W}$, ${Z}_\text{retained}=f_{sub}(\{{A}\odot{M_A},{X}\}, {W_{init}}\odot{M_W})$  as outlined Algorithm~\ref{alg:alg1}.
\STATE Compute ${M_A^r} = {A}\oplus{M_A}$ and ${M_W^r} = \lnot {{M_W}}$
\STATE Set ${Z}_\text{pruned}=f_{sub}(\{{A}\odot{M_A^r},{X}\}, {W_{init}}\odot{M_W^r})$
\FOR{round $t = 0, 1, 2, ..., T-1$}
\STATE Train ${Z}_\text{retained}$ and update ${M_A}$, ${M_W}$ according to $\mathcal{L}_{\text{retained}}$ computed in Equation~\eqref{eq2}.
\STATE Train ${Z}_\text{pruned}$ and update ${M_A^r}$, ${M_W^r}$ according to $\mathcal{L}_{\text{pruned}}$ computed in Equation~\eqref{eq3}.
\FOR{sample $k = 0, 1, 2, ..., \mathcal{K}-1$}
\STATE Sample ${u_\text{pruned}^{\omega}}$ and ${u_\text{pruned}^{\alpha}}$ in Equation~\eqref{eq4}
\STATE Sample ${u_\text{retained}^{\omega}}$ and ${u_\text{retained}^{\alpha}}$ in Equation~\eqref{eq5}
\ENDFOR
\STATE Remove the repeatedly sampled elements and obtain $\omega_\text{retained}$, $\omega_\text{pruned}$, $\alpha_\text{retained}$, $\alpha_\text{pruned}$.
\STATE Calculate the $\text{similarity}$ in Equation~\eqref{eq7}
\IF{$\text{similarity}>\text{threshold}$}
\STATE $\mathcal{K}\leftarrow\mathcal{K}/{2}$ 
\STATE Rewind masks to last round and re-sample.
\ENDIF
\STATE Exchange elements and obtain refined masks${M_A^{\prime}}$, ${M_W^{\prime}}$ according to Equation~\eqref{eq6}.
\ENDFOR
\end{algorithmic}
\end{algorithm}

\begin{algorithm}[!t]
\caption{Iterative ACE for GLT} 
\label{alg:alg3}
\begin{algorithmic}[1]
\REQUIRE GNN's initialization ${W_{init}}$, Pre-defined sparsity levels ${s_A}$ for graphs and ${s_W}$, Graph $\mathcal{G}=\{{A},{X}\}$, GNN $f(\cdot;{W_{init}})$, initial masks ${M_A} = A$, ${M_W}=1 \in \mathbb{R}^{\left\|{W_{init}}\right\|_0}$
\ENSURE GLT: ${Z}_{GLT}=f_{sub}(\{{A}\odot{M_A},{X}\};{W_{init}}\odot{M_W})$
\WHILE{$1-\frac{\left\|{M_A}\right\|_0}{\|{A}\|_0}<s_A \text { and } 1-\frac{\left\|{M_W}\right\|_0}{\|{W}\|_0}<s_W$}
\STATE Get refined masks ${M_A^{\prime}}$ and ${M_W^{\prime}}$ with GNN $f_{sub}(\cdot;{W}_{init}\odot{M_W})$ and graph $\mathcal{G}_{sub}=\{{A}\odot{M_A},X\}$ in Algorithm~\ref{alg:alg2}.
\STATE Update ${M_A}$=${M_A^{\prime}}$ and ${M_W}$=${M_W^{\prime}}$
\STATE Rewinding GNN weights to ${W}_{init}$.
\ENDWHILE
\end{algorithmic}
\end{algorithm}

After that, we make a refinement to the retained parts. For conciseness, we only introduce the refinement process of the model weights mask in the following, and the refinement process of the graph is similar. First, we use the magnitude to express the importance of the elements in ${M_W}$ and ${M_W^r}$.
Then, we use the gumbel-max method~\cite{jang2016categorical} to obtain the most discerning elements in ${M_W^r}$:
\begin{equation}
\label{eq4}
\begin{aligned}
u^{\omega}_{\text {pruned}} = \underset{u}{\arg\max }(\log {\left|{{M_{W}^r(u)}}\right|}+\epsilon_u).
\end{aligned}
\end{equation}
We define $\boldsymbol{v}$ as a one-hot vector whose dimension is equal to the number of elements in ${M_W^r}$, where $\boldsymbol{v}(u^{\omega}_{\text {pruned}})=1$ and others are 0. For more exploration, the term $\epsilon_u$ is randomly sampled from the Gumbel distribution and serves as a small amount of noise to avoid the $\arg\max$ operation always selecting the largest element. 
To excavate important elements as much as possible, we sample $\mathcal{K}$ times from the ${M}_{W}^r$ to obtain $\mathcal{K}$ most discerning elements, where the $\mathcal{K}$ is the upper limit of samples. Since duplicate sampling may occur, we follow the principle of keeping only one instance if an element is sampled more than once. As a consequence, the actual number of sampled elements $\mathcal{K}^{\prime}$ is often lower than $\mathcal{K}$.
We mark the sampled $\mathcal{K}^{\prime}$ elements as ${\omega}_\text{pruned}$, whose size is consistent with the size of ${M_W^r}$. If an element in ${M_W^r}$ is selected, the value of the corresponding position in $\omega_\text{retained}$ is 1, otherwise is 0. Similarly, we perform the same sampling operation to locate the most disturbing part of ${M_W}$: 
\begin{equation}
\label{eq5}
\begin{aligned}
u^{\omega}_{\text {retained}}= \underset{u}{\arg\max }(-\log {\left|{{M_{W}}(u)}\right|}+\epsilon_u),
\end{aligned}
\end{equation}
and mark the sampled elements of ${M_W}$ as ${\omega}_\text{retained}$, with identical size and assignment criteria to ${\omega}_\text{pruned}$. Notably, Equation~\ref{eq5} has a negative sign compared with Equation~\ref{eq4}, which means that the selection criteria is opposite.

Now, we can refine the original mask in ${Z}_\text{retained}$ by erasing the sampled elements of ${M_W}$ and adding the elements extracted from ${M_W^r}$ to ${M_W}$: 
\begin{align}
{M_W}^{\prime}&= {M_W} \oplus {\omega}_{\text{retained}} \oplus {\omega}_{\text{pruned}},\label{eq6}
\end{align}
where ${M_W}$ and ${M_W^r}$ are the original weight masks of ${Z}_\text{retained}$ and ${Z}_\text{pruned}$, $\oplus$ denotes the element-wise XOR (exclusive or) operator. 

Additionally, we observe a trade-off in choosing the value of $\mathcal{K}$. For fully excavating the pruned part, we would like to choose a large $\mathcal{K}$, which brings more computational costs. We further design an adaptive adjustment strategy for $\mathcal{K}$ to address this issue. First, we calculate the cosine similarity between different rounds:
\begin{equation}
\begin{aligned}
\label{eq7}
\text{similarity}  =\cos({\omega}_{\text{retained}}^{i+1}, {\omega}_{\text{pruned}}^{i})
 =\frac{\left({{\omega}_{\text{retained}}^{i+1}}^T {{\omega}_{\text{pruned}}^{i}} \right)}{||{\omega}_{\text{retained}}^{i+1}|| \cdot ||{\omega}_{\text{pruned}}^{i}|| },
\end{aligned}
\end{equation}
where ${\omega}_{\text{retained}}^{i+1}$ represents the elements sampled from the ${Z}_\text{retained}$ in round $i+1$ and ${\omega}_{\text{pruned}}^{i}$ represents the elements sampled from the ${Z}_\text{pruned}$ in round $i$. If $\text{similarity}$ is high, it means that most of the newly added elements in the $i\mbox{-}$th round will be erased in the following round, indicating that many unnecessary elements are sampled in the $i\mbox{-}$th round. Therefore, we halve the sampling limit and fallback to the state of the previous round to re-sample and swap. The refinement of adjacency matrix mask is similar to that of model weights. The detailed procedures of our ACE framework are summarized in Algorithm \ref{alg:alg2}.\\

\noindent\textbf{Discovering GLT with ACE. }The classical search algorithm of graph lottery tickets finds the winning tickets through an iterative magnitude-based pruning algorithm. 
Similarly, we employ the ACE framework to re-evaluate the importance of the pruned elements in the current state before further sparsification and extract useful information from the pruned parts to compensate for the performance loss. 
Then, rewind GNN's weights to the original initialization ${W}_{init}$. We repeat these processes until reaching pre-defined graph sparsity $s_A$ and model sparsity $s_W$, as shown in Algorithm~\ref{alg:alg3}.

\section{Experiments}
To demonstrate the effectiveness of the proposed ACE-GLT framework equipped with different GNNs, we carry out experiments on both node classification and link prediction tasks.
In this section, we first introduce the experimental settings in Section~\ref{section:4.1}, then provide the experimental results on the small-scale datasets and shallow GNNs in Section~\ref{section:4.2}, as well as the results on large-scale datasets and deeper GNNs in Section~\ref{section:4.3}. Furthermore, the ablation studies are given in Section~\ref{section:4.4}.

\subsection{Experimental Settings\label{section:4.1}}

\noindent\textbf{Datasets.} We first conduct experiments on three standard small-scale graph benchmark datasets to evaluate the effectiveness of ACE-GLT, including Cora, Citeseer, and PubMed~\cite{kipf2016semi}. Then, we further adopt the Open Graph Benchmark datasets~\cite{hu2020open} to verify the scalability of ACE-GLT on large-scale datasets, including Ogbn-ArXiv, Ogbn-Proteins, and Ogbl-Collab datasets. More details about the statistics of datasets are described in Appendix A. 

\noindent\textbf{Backbones.} To evaluate the generalizability of the ACE-GLT across different GNN backbones, we use three shallow network architectures as the backbones for small-scale datasets, including GCN~\cite{kipf2016semi}, GIN~\cite{xu2018powerful}, and GAT~\cite{velivckovic2017graph}. Moreover, we use a 28-layer ResGCN~\cite{li2020deepergcn} as the backbone for large-scale datasets. More details about the model parameters are described in Appendix B. 

\noindent\textbf{Metrics.} Considering the purpose of our work is to save the computational cost in inference by joint graph and model sparsification, we use maximum \emph{graph sparsity} and \emph{model sparsity} that can be achieved by the found GLT as primary metrics. The graph sparsity represents the ratio of the number of pruned edges to the number of edges in the original graph, and the model sparsity represents the ratios of the number of pruned weights to the number of weights in the original network. In addition, we also consider the \emph{highest accuracy} that can be achieved by sparse substructures in the process of iterative sparsification as a secondary metric.

\noindent\textbf{Tasks. } In this section we will focus on the experimental results on the node classification task under transductive setting. We also conduct experiments in the inductive setting and link prediction tasks, these results will be presented in Appendix D and Appendix E.

\subsection{Results on Small-Scale Datasets}
\label{section:4.2}
We first compare our method with UGS~\cite{chen2021unified}, GEBT~\cite{you2022early} and random pruning on small-scale datasets for node classification tasks. To better investigate how the accuracy varies with increasing model sparsity, we restore the mask of the adjacency matrix to its original state before further sparsification to ensure that the graph sparsity is always kept at a certain value~(5\%). Similarly, we fix the model sparsity at 20\% to explore the accuracy variation at different graph sparsity. Observing the results shown in Table~\ref{table1} and Figure S2 of Appendix, we can draw the following remarkable observations.

\begin{table*}[t]
\captionsetup{justification=centering}
\caption{Experimental results on node classification tasks of small-scale datasets~(Cora, Citeseer and PubMed). We choose GCN, GIN and GAT as backbones. \textbf{Bold} and \underline{underline} denote the best and the second-best results, respectively. $\pm$ corresponds to one standard deviation of the average evaluation over 3 trials. ``-'' means that GLT cannot be found. ``*'' implies statistical significance for $p < 0.05$ under paired t-test.}
\setlength\tabcolsep{0pt}
\renewcommand{\arraystretch}{1.6}
\resizebox{\linewidth}{!}{
\begin{tabular*}
{\linewidth}{@{}ccclclclclclclclclcl@{}}
\toprule
\multirow{2}{*}{\textbf{Backbone}\;} & \multirow{2}{*}{\;\textbf{Method}} & \multicolumn{6}{c}{\textbf{Model Sparsity}}                                                                                                            & \multicolumn{6}{c}{\textbf{Graph Sparsity}}                                                                                                            & \multicolumn{6}{c}{\textbf{Highest Accuracy}}                                                                                                          \\ \cmidrule(l){3-20} 
                           &                          & \multicolumn{2}{c}{\textbf{Cora}}                      & \multicolumn{2}{c}{\textbf{Citeseer}}                  & \multicolumn{2}{c}{\textbf{Pubmed}}                    & \multicolumn{2}{c}{\textbf{Cora}}                      & \multicolumn{2}{c}{\textbf{Citeseer}}                  & \multicolumn{2}{c}{\textbf{Pubmed}}                    & \multicolumn{2}{c}{\textbf{Cora}}                      & \multicolumn{2}{c}{\textbf{Citeseer}}                  & \multicolumn{2}{c}{\textbf{Pubmed}}                    \\ \midrule
\multirow{4}{*}{\textbf{GCN}}       & \textbf{Random}                  & \multicolumn{2}{c}{-}                         & \multicolumn{2}{c}{\underline{49.61 $\pm$ 0.28}}          & \multicolumn{2}{c}{-}                         & \multicolumn{2}{c}{-}                         & \multicolumn{2}{c}{-}                         & \multicolumn{2}{c}{-}                         & \multicolumn{2}{c}{82.05 $\pm$ 0.07}          & \multicolumn{2}{c}{71.00 $\pm$ 0.14}          & \multicolumn{2}{c}{79.15 $\pm$ 0.07}          \\
                           & \textbf{UGS}                     & \multicolumn{2}{c}{\underline{67.23 $\pm$ 0.23}}          & \multicolumn{2}{c}{36.14 $\pm$ 0.18}          & \multicolumn{2}{c}{\underline{95.63 $\pm$ 0.37}}          & \multicolumn{2}{c}{-}                         & \multicolumn{2}{c}{\underline{32.20 $\pm$ 0.24}}          & \multicolumn{2}{c}{11.75 $\pm$ 0.20}          & \multicolumn{2}{c}{\underline{82.12 $\pm$ 0.03}}          & \multicolumn{2}{c}{70.95 $\pm$ 0.21}          & \multicolumn{2}{c}{\underline{79.58 $\pm$ 0.04}}          \\
                           & \textbf{GEBT}            & \multicolumn{2}{c}{59.36 $\pm$ 0.27}          & \multicolumn{2}{c}{48.79 $\pm$ 0.27}          & \multicolumn{2}{c}{89.33 $\pm$ 0.30}          & \multicolumn{2}{c}{-}                         & \multicolumn{2}{c}{14.26 $\pm$ 0.10}          & \multicolumn{2}{c}{\underline{13.73 $\pm$ 0.04}}          & \multicolumn{2}{c}{81.98 $\pm$ 0.04}          & \multicolumn{2}{c}{\underline{71.18 $\pm$ 0.25}}          & \multicolumn{2}{c}{79.48 $\pm$ 0.03}          \\
                           & \textbf{Ours}                     & \multicolumn{2}{c}{\;\textbf{89.64 $\pm$ 0.45$^*$}} & \multicolumn{2}{c}{\;\textbf{83.91 $\pm$ 0.18$^*$}} & \multicolumn{2}{c}{\;\textbf{97.21 $\pm$ 0.07$^*$}} & \multicolumn{2}{c}{\textbf{9.84 $\pm$ 0.35}}  & \multicolumn{2}{c}{\;\textbf{40.42 $\pm$ 0.18$^*$}} & \multicolumn{2}{c}{\;\textbf{22.66 $\pm$ 0.45$^*$}} & \multicolumn{2}{c}{\textbf{82.23 $\pm$ 0.11}} & \multicolumn{2}{c}{\textbf{71.53 $\pm$ 0.18}} & \multicolumn{2}{c}{\;\textbf{80.48 $\pm$ 0.18$^*$}} \\ \midrule
\multirow{4}{*}{\textbf{GIN}}       & \textbf{Random}                   & \multicolumn{2}{c}{-}                         & \multicolumn{2}{c}{60.08 $\pm$ 0.16}          & \multicolumn{2}{c}{74.19 $\pm$ 0.16}          & \multicolumn{2}{c}{13.74 $\pm$ 0.06}          & \multicolumn{2}{c}{25.31 $\pm$ 0.27}          & \multicolumn{2}{c}{-}                         & \multicolumn{2}{c}{78.88 $\pm$ 0.11}          & \multicolumn{2}{c}{68.93 $\pm$ 0.18}          & \multicolumn{2}{c}{78.83 $\pm$ 0.11}          \\
                           & \textbf{UGS}                      & \multicolumn{2}{c}{90.96 $\pm$ 0.40}          & \multicolumn{2}{c}{\underline{96.49 $\pm$ 0.03}}          & \multicolumn{2}{c}{\underline{92.32 $\pm$ 0.24}}          & \multicolumn{2}{c}{\underline{24.36 $\pm$ 0.27}}          & \multicolumn{2}{c}{26.48 $\pm$ 0.21}          & \multicolumn{2}{c}{\underline{15.27 $\pm$ 0.41}}          & \multicolumn{2}{c}{\underline{79.48 $\pm$ 0.04}}          & \multicolumn{2}{c}{\underline{69.83 $\pm$ 0.11}}          & \multicolumn{2}{c}{79.20 $\pm$ 0.14}          \\
                           & \textbf{GEBT}                     & \multicolumn{2}{c}{\underline{92.94 $\pm$ 0.51}}          & \multicolumn{2}{c}{95.00 $\pm$ 0.78}          & \multicolumn{2}{c}{86.00 $\pm$ 0.96}          & \multicolumn{2}{c}{18.57 $\pm$ 0.13}          & \multicolumn{2}{c}{\underline{30.15 $\pm$ 0.20}}          & \multicolumn{2}{c}{12.26 $\pm$ 0.20}          & \multicolumn{2}{c}{78.95 $\pm$ 0.21}          & \multicolumn{2}{c}{69.72 $\pm$ 0.11}          & \multicolumn{2}{c}{\underline{79.39 $\pm$ 0.13}}          \\
                           & \textbf{Ours}                     & \multicolumn{2}{c}{\;\textbf{94.38 $\pm$ 0.25$^*$}} & \multicolumn{2}{c}{\;\textbf{98.56 $\pm$ 0.20$^*$}} & \multicolumn{2}{c}{\;\textbf{95.13 $\pm$ 0.65$^*$}} & \multicolumn{2}{c}{\;\textbf{37.14 $\pm$ 0.35$^*$}} & \multicolumn{2}{c}{\;\textbf{36.97 $\pm$ 0.23$^*$}} & \multicolumn{2}{c}{\;\textbf{30.20 $\pm$ 0.47$^*$}} & \multicolumn{2}{c}{\;\textbf{80.00 $\pm$ 0.14$^*$}} & \multicolumn{2}{c}{\;\textbf{70.23 $\pm$ 0.18$^*$}} & \multicolumn{2}{c}{\textbf{79.48 $\pm$ 0.11}} \\ \midrule
\multirow{4}{*}{\textbf{GAT}}       & \textbf{Random}                   & \multicolumn{2}{c}{73.37 $\pm$ 0.42}          & \multicolumn{2}{c}{80.01 $\pm$ 2.36}          & \multicolumn{2}{c}{-}                         & \multicolumn{2}{c}{\;9.36 $\pm$ 0.16}           & \multicolumn{2}{c}{-}                         & \multicolumn{2}{c}{-}                         & \multicolumn{2}{c}{78.15 $\pm$ 0.21}          & \multicolumn{2}{c}{67.25 $\pm$ 0.35}          & \multicolumn{2}{c}{77.28 $\pm$ 0.04}          \\
                           & \textbf{UGS}                      & \multicolumn{2}{c}{\underline{91.71 $\pm$ 0.45}}          & \multicolumn{2}{c}{\underline{92.44 $\pm$ 1.60}}          & \multicolumn{2}{c}{\underline{95.63 $\pm$ 0.45}}          & \multicolumn{2}{c}{37.14 $\pm$ 1.61}          & \multicolumn{2}{c}{\underline{40.11 $\pm$ 0.13}}          & \multicolumn{2}{c}{\underline{54.01 $\pm$ 0.34}}          & \multicolumn{2}{c}{\underline{79.05 $\pm$ 0.07}}          & \multicolumn{2}{c}{\underline{68.33 $\pm$ 0.11}}          & \multicolumn{2}{c}{78.05 $\pm$ 0.07}          \\
                           & \textbf{GEBT}                     & \multicolumn{2}{c}{84.17 $\pm$ 0.33}          & \multicolumn{2}{c}{86.62 $\pm$ 1.77}          & \multicolumn{2}{c}{94.54 $\pm$ 0.61}         & \multicolumn{2}{c}{\underline{40.40 $\pm$ 0.23}}          & \multicolumn{2}{c}{14.26 $\pm$ 1.60}          & \multicolumn{2}{c}{46.31 $\pm$ 0.14}          & \multicolumn{2}{c}{78.90 $\pm$ 0.14}          & \multicolumn{2}{c}{68.27 $\pm$ 0.10}          & \multicolumn{2}{c}{\underline{78.13 $\pm$ 0.18}}          \\
                           & \textbf{Ours}                     & \multicolumn{2}{c}{\;\textbf{93.37 $\pm$ 0.30$^*$}} & \multicolumn{2}{c}{\textbf{93.15 $\pm$ 0.20}} & \multicolumn{2}{c}{\;\textbf{97.76 $\pm$ 0.68$^*$}} & \multicolumn{2}{c}{\;\textbf{46.31 $\pm$ 1.57$^*$}} & \multicolumn{2}{c}{\textbf{\;49.72 $\pm$ 1.10$^*$}} & \multicolumn{2}{c}{\;\textbf{58.58 $\pm$ 0.37$^*$}} & \multicolumn{2}{c}{\textbf{79.18 $\pm$ 0.11}} & \multicolumn{2}{c}{\textbf{68.49 $\pm$ 0.13}} & \multicolumn{2}{c}{\textbf{78.38 $\pm$ 0.11}} \\ \bottomrule
\end{tabular*}}
\label{table1}
\end{table*}

\noindent\textbf{Observation 1: ACE-GLT can find GLTs with sparser subgraphs/subnetworks.} ACE-GLT consistently outperforms other methods under the same experimental setting across all datasets and GNNs, which validates the superiority of refining the search process with pruned elements. For example, the graph lottery ticket on Citeseer+GCN identified by ACE-GLT achieves 83.91\% model sparsity, while UGS can only achieve 36.14\%. In addition, ACE-GLT can find the graph lottery ticket with 37.14\% graph sparsity on Cora+GIN, which outperforms UGS by 13\%. Overall, our method makes $5\% \sim 10\%$ graph sparsity improvement and $2\% \sim 47\%$ model sparsity improvement for graph lottery tickets.

\noindent\textbf{Observation 2: ACE-GLT can solve the cases in which previous methods fail to find GLT.} For the node classification task on Cora+GCN, previous works fail to find graph lottery ticket with sparse graph when model sparsity is fixed at 20\%. Apparently, ACE-GLT can find a graph lottery ticket with 9.84\% graph sparsity and 20\% model sparsity for the first time, making a impressive breakthrough in this case.

\noindent\textbf{Observation 3: The sensitivity to the refinement depends on the properties of the input graphs and backbones.} First, ACE-GLT improves more significantly on smaller datasets. We find that Cora and Citeseer are more sensitive to pruning, while PubMed is more robust to sparsification, such as 47\% model sparsity improvement on Citeseer+GCN and only 2\% model sparsity improvement on PubMed+GCN. We argue that the graph property such as graph size may be a crucial factor to this result. Smaller graphs are more vulnerable to pruning operations and more likely to lose valid information, while ACE-GLT can obtain a more informative substructure and thus effectively remedy the information loss, making the performance of graph lottery tickets found by ACE-GLT on smaller datasets far ahead. Apart from that, we notice GCN and GIN are more sensitive to refinement operations than GAT. A possible explanation is that the attention-based aggregation mechanism adopted by GAT is capable of re-identifying important connections in sparse graphs, allowing GAT to recognize truly important parts even without refinement.

\noindent\textbf{Observation 4: ACE-GLT can obtain sparse substructures with higher accuracy.}
In addition to the highest sparsity that can be achieved, we also observe the prediction accuracy of sparse substructures obtained in progressive pruning. The accuracy variation of the substructures obtained by each method during the gradual sparsification process is shown in Figure S2 of Appendix C. We find that when achieve same sparsity, substructures refined by our method can achieve higher accuracy compared to those that are not refined. For example, for the node classification task on Cora+GCN, when the model sparsity reaches 74\%, the accuracy of the substructure corrected by ACE-GLT can reach 82.3, which is 0.5 higher compared to other methods. And when the model sparsity increases to 95\%, the difference in accuracy even increases to 2.6.

\noindent\textbf{Observation 5: ACE-GLT is more amenable to extreme sparsity.} 
As shown in Table~\ref{table2}, when the model sparsity reached 90\%, the model shows no significant performance degradation and even exceeds the original performance on most of the datasets and backbones. Moreover, ACE-GLT makes a more significant improvement on the accuracy of the sparser substructures.

We will represent more experimental results under different fixed graph/model sparsity in Appendix F. 

\subsection{Results on Large-Scale Datasets}
\label{section:4.3}
\begin{figure}[!ht]
\centering
\includegraphics[width=1.0\linewidth]{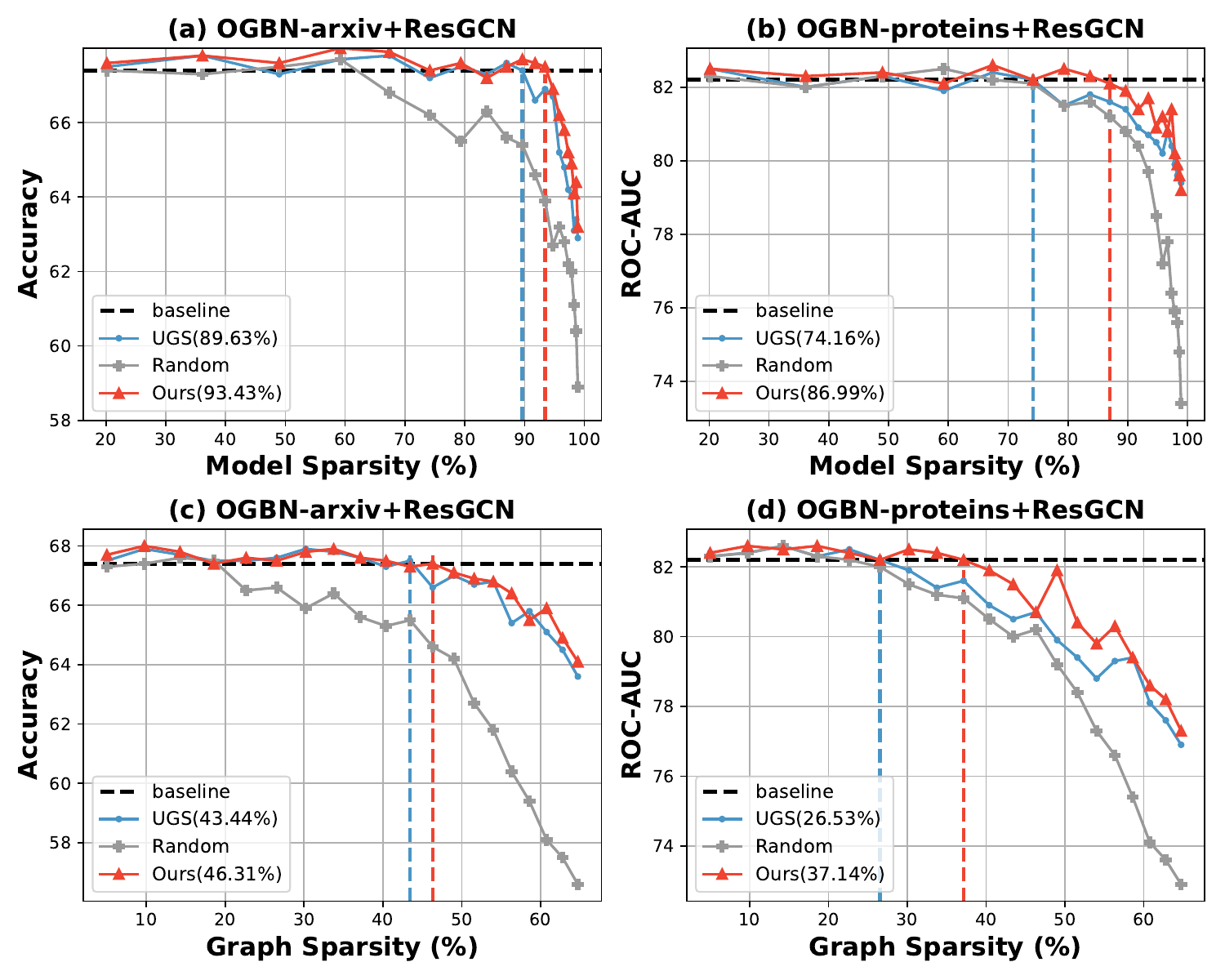}
\captionsetup{justification=centering}
\caption{Experimental results on node classification tasks of large-scale datasets (OGBN-arxiv and OGBN-proteins). We choose deep graph neural networks (28-layer deep ResGCN) as backbone. The performance of our framework is compared with UGS and random pruning. The ``vertical dotted lines'' with different colors in each figure  represent the maximum model/graph sparsity of GLTs found by different methods. }
\label{fig:fig3}
\end{figure}

To further investigate whether our approach can still achieve excellent results on large-scale datasets and deeper graph neural networks, we conduct experiments on ResGCN with the OGBN-arxiv, OGBN-proteins and OGBL-Collab. The first two datasets are used for the node classification task, while the third one is for the link prediction task. We still set graph/model sparsity to a fixed value and gradually increase model/graph sparsity. The results are shown in Figure~\ref{fig:fig3}, and we can draw another observation.

\noindent\textbf{Observation 6: ACE-GLT performs equally well on large-scale datasets.} First, ACE-GLT can find graph lottery tickets with sparser graphs and models on both OGBN-arxiv and OGBN-proteins. On the OGBN-proteins dataset, ACE-GLT obtains graph lottery ticket with 86.99\% model sparsity and 37.14\% graph sparsity, while other methods can only achieve 74.16\% model sparsity and 26.53\% graph sparsity. The same holds for the OGBN-arxiv dataset, where we improve the model sparsity and graph sparsity of the winning tickets by 5\% and 4\%, respectively. Such improvement has great significance for OGB datasets with millions of edges and ResGCN with an enormous number of parameters. Secondly, in the process of progressive sparsification, the obtained sparse models and graphs~(not necessarily GLTs) refined by ACE-GLT achieve higher accuracy than those pruned by the original methods.

\subsection{Ablation Study}

\label{section:4.4}
\begin{figure}[!b]
\centering
\includegraphics[width=1.0\linewidth]{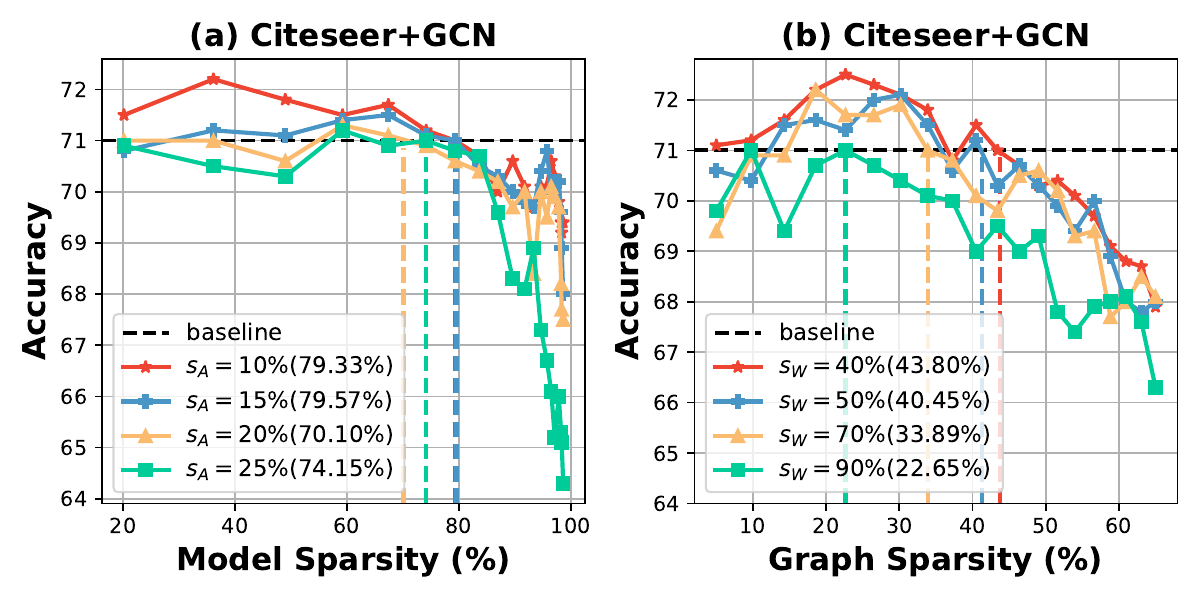}
\captionsetup{justification=centering}
\caption{Experimental results on Citeseer+GCN with different graph sparsity or model sparsity. (a) This is the curve of accuracy with increasing model sparsity for different fixed graph sparsity. (b) This is the curve of accuracy with increasing graph sparsity for different fixed model sparsity.}
\label{fig:fig4}
\end{figure}

\noindent\textbf{Pruning ratio ${p_A}$ and ${p_W}$.}
We further complete the experiments in Section 4.2 by investigating the pruning ratios $p_A$ and $p_W$ in ACE-GLT for graphs and model sparsification. We adjust different fixed graph sparsity~(${p_A}=10\%, 15\%, 20\%, 25\%$) and model sparsity ~(${p_W}=40\%, 50\%, 70\%, 90\%$) to examine the effect of different sparsity settings on the performance of proposed ACE-GLT. Figure~\ref{fig:fig4} reports the experimental results on Citeseer+GCN. First, it can be easily observed that our method can obtain similar results under different graph sparsity settings~(shown in Figure~\ref{fig:fig4}(a)). For example, given ${p_A}=10\%$, it is possible to obtain a subnetwork with 79.33\% model sparsity while maintaining accuracy, which is only 5\% higher than the subnetwork obtained when setting ${p_A}=25\%$. In addition, we note that increasing model sparsity ${p_W}$ may lead to significant accuracy degradation. Given ${p_W}=70\%$, ACE-GLT can identify an subgraph with 43.80\% sparsity and ensure that the performance will not decrease. However, ACE-GLT only identifies such subgraph with 22.65\% sparsity when ${p_W}$ reaches 90\%. Therefore, we suppose that the input graph tends to be more sensitive to refinement operation than the model parameters.\\

\begin{table}[!t]
\huge
\renewcommand{\arraystretch}{1.6}
\resizebox{\columnwidth}{!}{
\begin{tabular}{@{}lcclcccccc@{}}
\toprule
\multicolumn{1}{c}{\textbf{}}& \multicolumn{3}{c}{\textbf{Setting}}& \multicolumn{6}{c}{\textbf{Model Sparsity}}\\ \cmidrule(l){2-10}
\textbf{}& \textbf{re-sample} & \multicolumn{2}{c}{\textbf{adaptive $\mathcal{K}$}} & \multicolumn{1}{l}{\textbf{20\%}}&\multicolumn{1}{l}{\textbf{40\%}} & \multicolumn{1}{l}{\textbf{50\%}} & \multicolumn{1}{l}{\textbf{60\%}} & \multicolumn{1}{l}{\textbf{70\%}} & \multicolumn{1}{l}{\textbf{80\%}} \\ \midrule
\multicolumn{1}{c}{\multirow{4}{*}{\textbf{Cora+GCN}}}& $\times$& \multicolumn{2}{c}{$\times$}& 82.2& 81.8& 82.1& 82.0& 81.9& \textbf{81.8}\\
\multicolumn{1}{c}{}& $\checkmark$& \multicolumn{2}{c}{$\times$}& 82.3& 82.2& 82.4& 82.2& 81.8& 81.2\\
\multicolumn{1}{c}{}& $\times$& \multicolumn{2}{c}{$\checkmark$}& 82.4& 82.0& 82.3& 81.9& 82.1& 81.5\\
\multicolumn{1}{c}{}& $\checkmark$& \multicolumn{2}{c}{$\checkmark$}& \textbf{82.6}& \textbf{82.5}& \textbf{83.1}& \textbf{82.7}&\textbf{82.2}& 81.7\\ \midrule
\multicolumn{1}{c}{\multirow{4}{*}{\textbf{Citeseer+GIN}}}& $\times$& \multicolumn{2}{c}{$\times$}& 68.3& 68.7& 68.7& 69.0& 68.7& 68.5\\ 
\multicolumn{1}{c}{}& $\checkmark$& \multicolumn{2}{c}{$\times$}& 68.7& 68.5& 68.9& 68.6& 69.0& 68.8\\
\multicolumn{1}{c}{}& $\times$& \multicolumn{2}{c}{$\checkmark$}& 68.5& 68.8& 68.7& \textbf{69.2}& 68.7& 68.5\\
\multicolumn{1}{c}{}& $\checkmark$& \multicolumn{2}{c}{$\checkmark$}& \textbf{68.7}& \textbf{68.9}& \textbf{69.5}& 68.5& \textbf{69.6}& \textbf{69.8}\\ \bottomrule 
\end{tabular}}
\captionsetup{justification=centering}
\caption{Ablation studies on re-sample and adaptive $\mathcal{K}$ techniques. We represent experiments on Cora+GCN and Citeseer+GIN here, and more results are included in Appendix F.}
\label{table2}
\end{table}

\noindent\textbf{With or without re-sample and adaptive $\mathcal{K}$ techniques.}
We adopt re-sample and adaptive $\mathcal{K}$ techniques in our framework to control the extent of the refinement operation and more accurately locate the elements that need to be removed or retrieved. To explore the effect of these techniques, we implement three variants: $1)$ with neither of them, $2)$ with re-sample technique~(without adaptive $\mathcal{K}$), and $3)$ with adaptive $\mathcal{K}$~(without re-sample technique) and compare their results with our framework. As depicted in table~\ref{table2}, we can observe that either re-sample or adaptive $\mathcal{K}$ can help boost performance, and the combination of these two techniques~(the solution adopted by our framework) has the best performance. Such merits stem from the fact that these techniques prevent genuinely valuable information from being replaced, effectively guaranteeing that the information exchanged between the retained and pruned substructures is exhaustive and non-redundant.

\section{Conclusion}
In this paper, we propose an effective method termed as adversarial complementary erasing for graph lottery tickets~(ACE-GLT), which extracts valid information from the pruned parts to improve the search of GLTs. First, Our work puts forward a conjecture that the dynamic changes in graph and model structure during iterative pruning may result in inaccurate evaluation of importance, leading to erroneous pruning and information loss. Then we propose an adversarial complementary erasing~(ACE) framework to replace the impurities in the retained part by continuously extracting effective elements from the pruned part, thus realizing the compensation for information loss. By iteratively applying ACE in gradual sparsification, we can refine the searching process and thus identify better-performing GLTs. We further apply our method to different tasks, backbones and datasets. Extensive experimental results demonstrate the effectiveness of our ACE-GLT, achieving $5\% \sim 10\%$ graph sparsity improvement and $2\% \sim 47\%$ model sparsity improvement for GLTs. These results also prove the existence of genuinely valid information in the pruned parts and verify our conjecture. 

\textbf{Limitations.} 
Our algorithm takes longer training time~(by about 2x) than other methods when searching for graph lottery tickets due to its re-evaluating and re-selecting mechanism.
However, it is more noteworthy that the graph lottery tickets found by our algorithm can achieve much higher sparsity, which will greatly reduce the time consuming in inference.
Considering the significant saving in inference time, the extra time consumption is worthwhile in real-world application where the training time is less concerned than the inference time. In addition, our work currently only applies to single-task learning on homogeneous graphs. In our future work, we will generalize our method to more complex scenarios, such as heterogeneous graphs and multi-task learning, to boost their efficiency.
\newpage
\section*{Acknowledgement}
This work is supported by the Science and Technology Project of SGCC: Hybrid Enhancement Intelligence with Human-AI Coordination and its Application in Reliability Analysis of Regional Power System (5700-202217190A-1-1-ZN).

\bibliography{ecai}

\newpage
\appendix
\setcounter{figure}{0}
\setcounter{table}{0}
\renewcommand\thefigure{S\arabic{figure}}
\renewcommand\thetable{S\arabic{table}}
\section{Importance Fluctuation}
\begin{figure*}[!ht]
  \centering
  \includegraphics[width=1.0\linewidth]{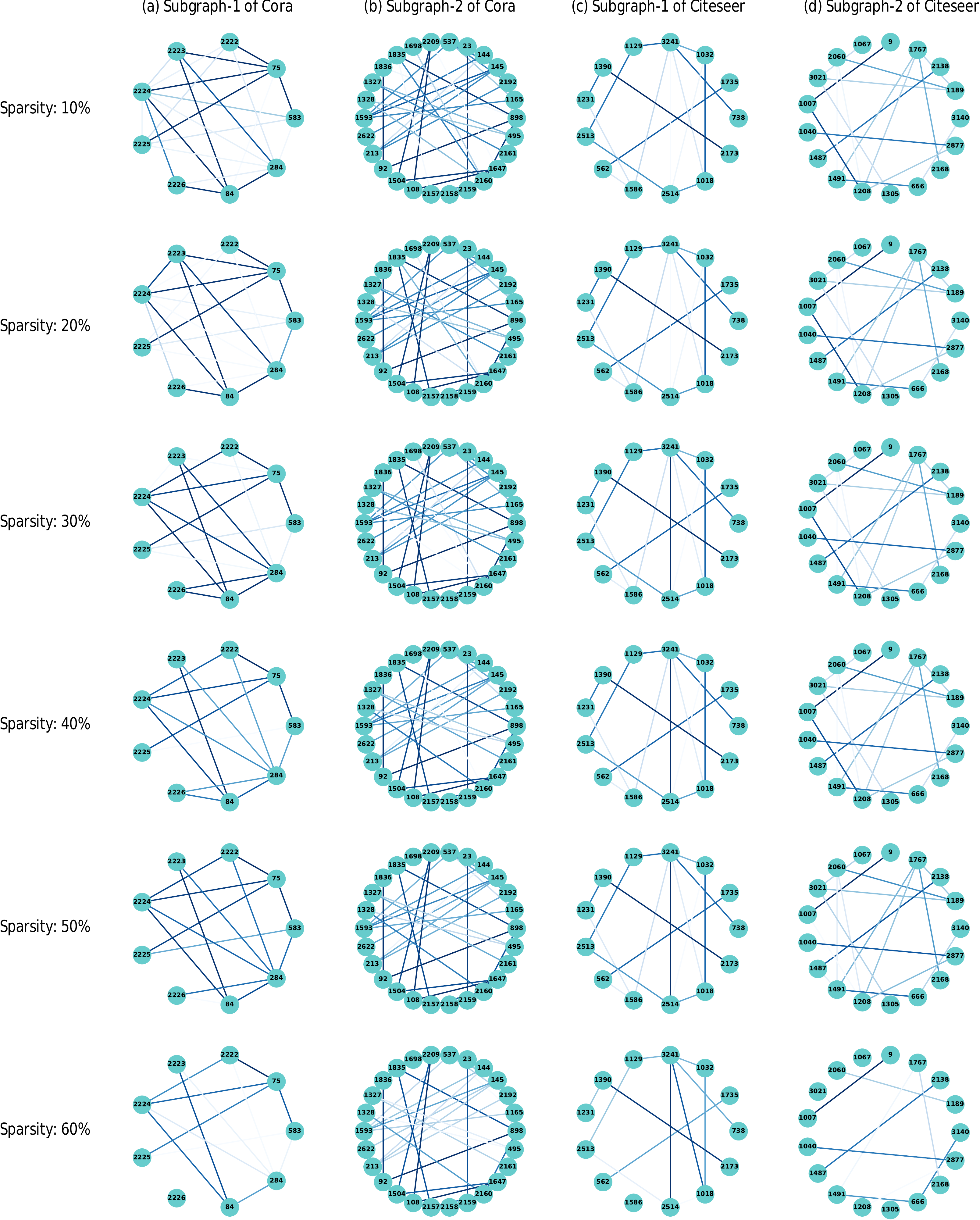}
  \captionsetup{justification=centering}
  \caption{Visualization of the edges importance at different stages of the iterative magnitude-based pruning process.}
  \label{fig:figs1}
\end{figure*}

We use the subgraph visualization in this section to better illustrate the problem of potential fluctuations in the importance ranking of elements during iterative magnitude-based pruning. As shown in Figure~\ref{fig:figs1}, we show the variation of the importance of edges for the connected subgraphs in the Cora and Citeseer dataset at two different sparsities, respectively. We use different colors to indicate importance levels of the edges, with darker colors indicating that the edges are more important.
Taking Figure~\ref{fig:figs1}(a) as an example, in the early iterations (sparsity = 10\%), the edge between node 2284 and node 284 is shown with darker color, indicating its high importance. However, after several rounds of pruning, in the later iterations (sparsity = 60\%), the same edge becomes much lighter in color, indicating its decreased importance. Additionally, the edge between node 2283 and node 284 shows an opposite trend, going from unimportant to important. Therefore, we conclude that the dynamic changes in the network structure can have a significant impact on the importance ranking of edges.

\section{More Implementation Details}
In this part, we first present relevant information about the datasets, including the detailed statistics of the datasets and the selected metrics.
In addition, we describe the structure of backbones and various hyper-parameters used for training.

\begin{table*}[b]
  \captionsetup{justification=centering}
  \caption{The detailed statistics of the datasets and their evaluation metrics.}
  \resizebox{\linewidth}{!}{
  \begin{tabular}{@{}cccccc@{}}
  \toprule
  \textbf{Dataset}      & \textbf{Task}                               & \textbf{\# Nodes}   & \textbf{\# Features}     & \textbf{\# Edges}      & \textbf{\# Classes} \\ \midrule
  Cora          & Node Classification \& Link Prediction & 2,708   & 1,433        & 5,429      & 7   \\
  Citeseer      & Node Classification \& Link Prediction & 3,327   & 3,703        & 4,732      & 6   \\
  PubMed        & Node Classification \& Link Prediction & 19,717  & 500          & 88,338     & 3   \\
  ogbn-arxiv    & Node Classification                 & 169,343 & 128          & 1,166,243  & 40  \\
  ogbn-proteins & Node Classification                 & 132,534 & 8 (on edges) & 39,561,252 & 112  \\
  ogbl-collab   & Link Prediction                     & 235,868 & 128          & 1,285,465  & 2  \\ 
  \bottomrule
  \end{tabular}}
  \label{tables1}
\end{table*}

\subsection{Datasets}
For small-scale datasets, we use three commonly used citation network datasets for node classification tasks and link prediction tasks, including Cora, Citeseer and PubMed.
In these datasets, each node represents a paper, the node's label represents the category of the paper, and the edges represent the citation relationships among these papers.
For large-scale datasets, we use three Open Graph Benchmark~(OGB) datasets, including ogbn-arxiv~(Paper Citation Network), ogbn-proteins~(Protein-Protein Association Network) and ogbl-collab~(Author Collaboration Network).
Ogbn-arxiv and ogbn-proteins are used for node classification tasks, while ogbl-collab is used for link prediction tasks.
The statistics of these datasets are listed in Table~\ref{tables1}.

\subsection{Train-val-test Splitting of Datasets}
To more fairly and convincingly verify the advantages of our method, we follows the datasets splitting criterion used by UGS~\cite{chen2021unified}.
For node classification tasks on small-scale datasets, we use 140~(Cora), 120~(Citeseer), and 60~(PubMed) labeled nodes for training, 500 nodes for validation and 1000 nodes for test.
And for link prediction, we shuffle the datasets and sample 85\% edges for training, 5\% for validation and 10\% for test.
For OGB datasets, we use the classification criteria given by~\cite{hu2020open}.
Specifically, ogbn-arxiv is split by year, and we use the papers published before 2017 for training, the papers published in 2018 for validation and the papers published in 2019 for test.
As for ogbn-proteins, we split the protein nodes into training/validation/test sets according to the species.
For ogbl-collab used in link prediction task, we adopt the collaborations until 2017 as training edges, those in 2018 as validation edges, and those in 2019 as test edges.

\subsection{Metrics}
In our experiments, we mainly use three different evaluation metrics for different tasks and datasets, including Accuracy, ROC-AUC~(Receiver Operating Characteristic-Area Under the Curve) and Hits@50.
Accuracy is the ratio of the number of correctly classified predictions to the total number of predictions, while higher accuracy means better model performance.
The value of ROC-AUC represents the probability that the model will score the given positive sample higher than the given negative sample, which is often used to evaluate the merit of a binary classifier.
And Hits@50 evaluates how well the model ranks the positive samples against the negative samples. Specifically, we rank all positive edges in the test dataset and a large number of negative edges from random sampling, then we calculate the proportion of positive edges ranked in top 50 or higher.
For different datasets and tasks, different metrics have been chosen to judge performance, and the selection of metrics for the OGB dataset was strictly based on~\cite{hu2020open}. The specific metrics for the different tasks and datasets are shown in Table~\ref{tables1}.

\subsection{Backbones}
For Cora, Citeseer and PubMed, we consistently use GCN/GIN/GAT with two layers as backbones and set the hidden units of them at 512. 
Additionally, for GAT, we use 8 heads in the first layer, and 1 head in the second layer.
Apart from them, we adopt 28-layer ResGCNs for OGB datasets.
\subsection{Training Details and Hyper-parameter Settings}
We compare our method with UGS~\cite{chen2021unified}, GEBT~\cite{you2022early} and random pruning, and follow the same parameter settings with them.
The hyper-parameters used in our approach include the number of training rounds in the pruning phase, the number of training rounds for the retained and pruned substructures in the refinement phase, the $\lambda_1$ and $\lambda_2$  for controlling regularity terms, the learning rate, the optimizer, etc. 
The specific settings are presented in Table~\ref{tables2}.
\begin{table*}[ht]
  \captionsetup{justification=centering}
  \caption{The details of hyper-parameters settings.}
  \resizebox{\linewidth}{!}{
  \begin{tabular}{@{}cccccccccc@{}}
  \toprule
                           & \multicolumn{5}{c}{Node Classification}                 & \multicolumn{4}{c}{Link Prediction}      \\ \cmidrule(l){2-10} 
                           & Cora   & Citeseer & PubMed & ogbn-arxiv & ogbn-proteins & Cora   & Citeseer & PubMed & ogbl-collab \\ \midrule
  Epochs~(pruning/refining) & 200/30 & 200/30   & 200/30 & 200/30     & 200/30        & 200/30 & 200/30   & 200/30 & 200/30      \\
  Optimizer                & Adam   & Adam     & Adam   & Adam       & Adam          & Adam   & Adam     & Adam   & Adam        \\
  Weight Decay             & 6e-5   & 5e-4     & 5e-4   & 0          & 0             & 0      & 0        & 0      & 0           \\
  $\lambda_1$              & 2e-3   & 1e-6     & 1e-2   & 1e-6       & 1e-1          & 1e-4   & 2e-3     & 1e-5   & 1e-6        \\
  $\lambda_2$              & 2e-3   & 1e-4     & 1e-2   & 2e-5       & 1e-2          & 1e-4   & 2e-3     & 1e-3   & 1e-5        \\
  Learning Rate            & 6e-2   & 1e-2     & 1e-2   & 1e-2       & 1e-2          & 3e-4   & 3e-4     & 3e-4   & 1e-2        \\ \bottomrule
  \end{tabular}}
  \label{tables2}
\end{table*}

\section{Accuracy Variation}
\begin{figure*}[!ht]
  \centering
  \includegraphics[width=1.0\linewidth, height=0.92\textheight]{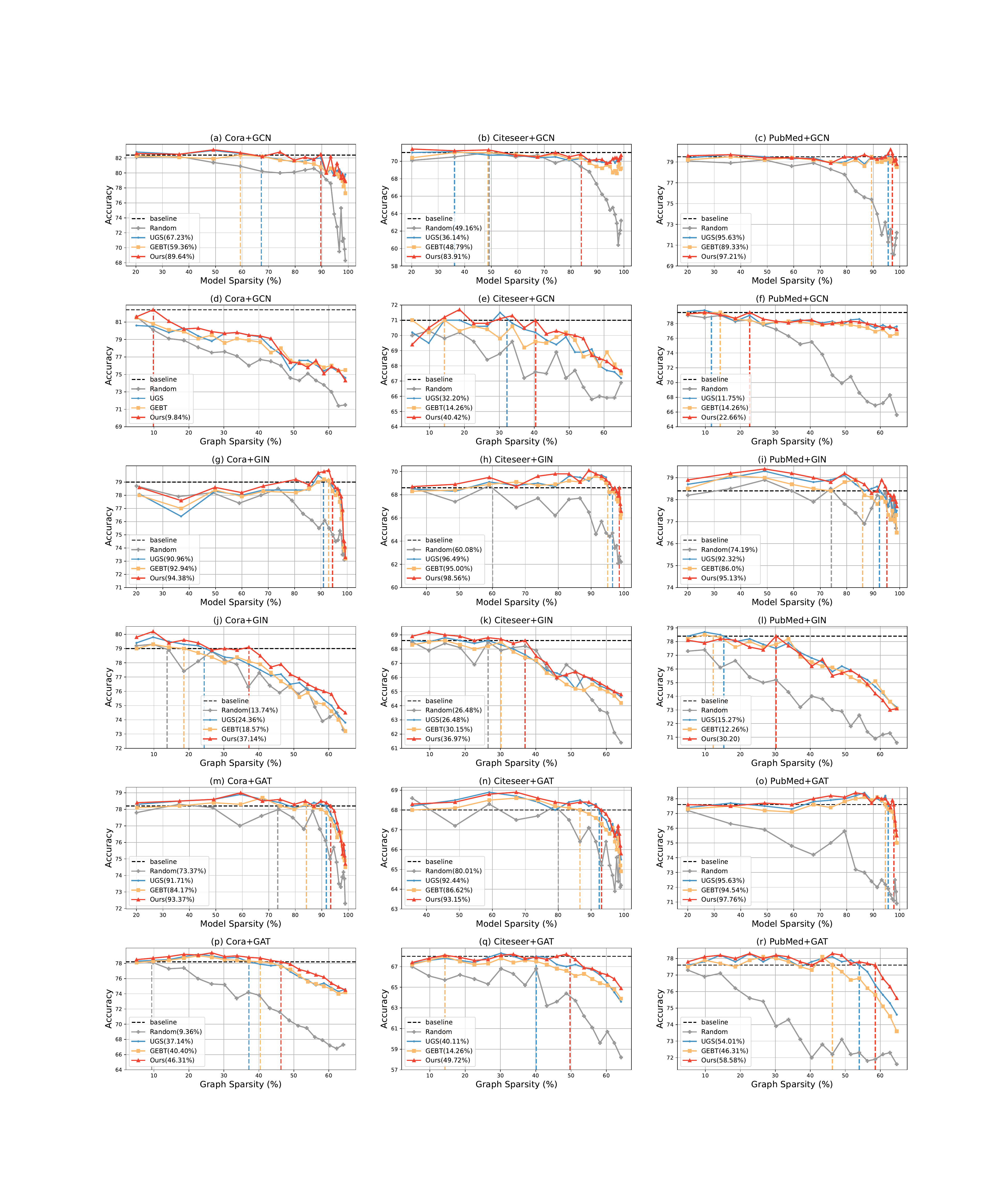}
  \captionsetup{justification=centering}
  \caption{Experimental results on node classification tasks of small-scale datasets~(Cora, Citeseer and PubMed). We choose GCN, GIN and GAT as backbones. The baseline represents the performance of full GNNs on full graphs. The vertical dotted lines with different colors indicate the model/graph sparsity of the GLTs identified by different methods.}
  \label{fig:figs2}
\end{figure*}

 In this part, we present the accuracy variation of experiments conducted on the Cora, Citeseer and PubMed datasets with GCN, GIN and GAT. As shown in Figure~\ref{fig:figs2}, the baseline represents the performance of full GNNs on full graphs. The ``vertical dotted lines'' with different colors indicate the highest model/graph sparsity of sparse substructures identified by different methods.

\section{Experiment Results on Link Prediction Tasks}
In this part, we conduct extensive experiments on several datasets for the link prediction tasks, including small and large scale datasets.
\subsection{Small-scale Datasets}

\begin{figure*}[!ht]
  \centering
  \includegraphics[width=1.0\linewidth,height=0.92\textheight]{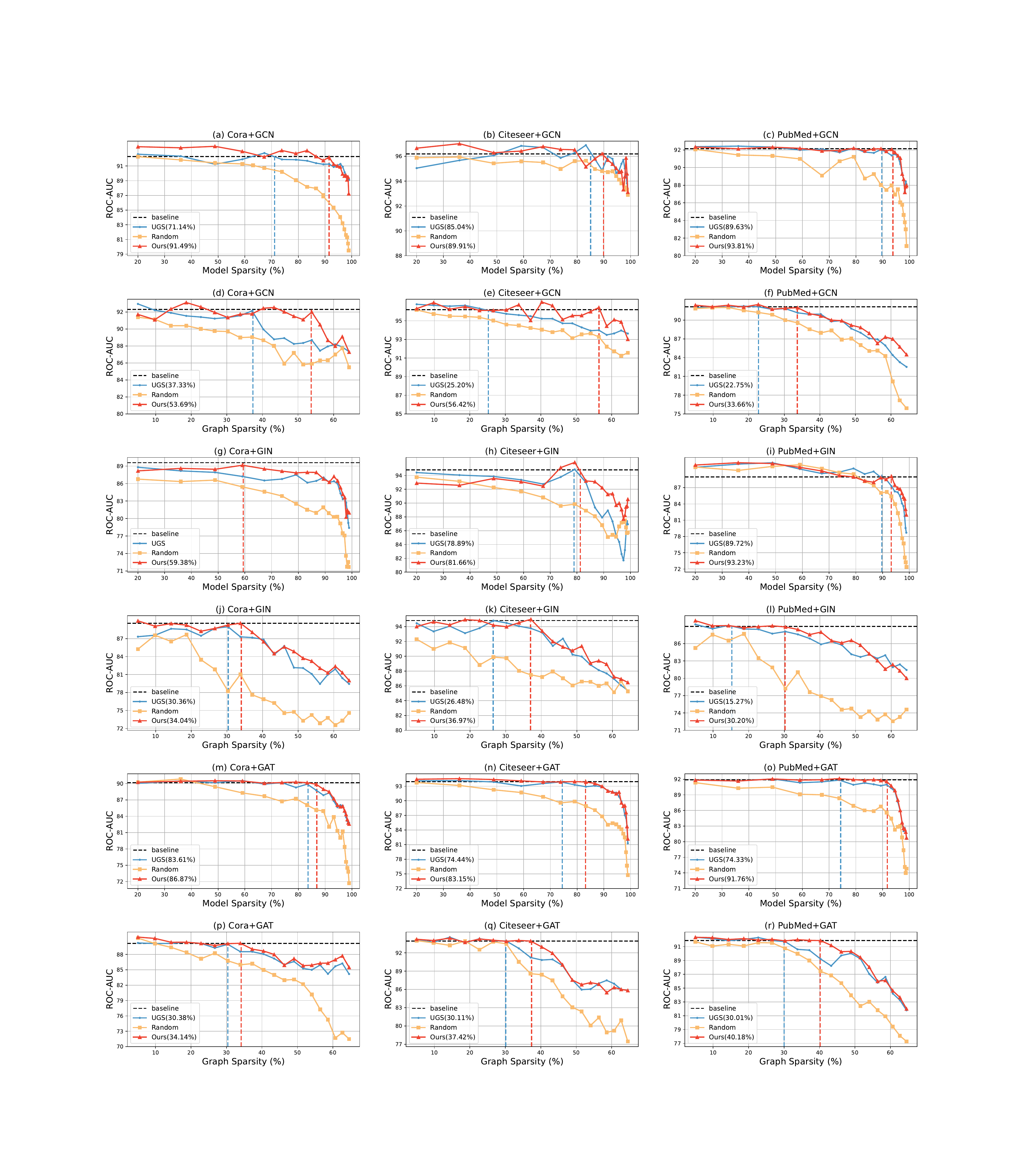}
  \captionsetup{justification=centering}
  \caption{Experimental results on link prediction tasks of small-scale datasets~(Cora, Citeseer and PubMed). We choose GCN, GIN and GAT as backbones.}
  \label{fig:figs3}
\end{figure*}

As shown in Table~\ref{tables3}, we present experiment results of link prediction with GCN/GIN/GAT on small-scale datasets, including Cora, Citeseer and PubMed.
We can obtain a similar conclusion to the node classification task: with our ACE-GLT, we can find GLTs with sparser subnetworks/subgraphs, and the retained substructures can obtain higher ROC-AUC after refinement under different sparsity levels.
Overall, with ACE-GLT used for refining the searching process, the identified GLTs can achieve $30\%\sim57\%$ graph sparsity and $60\%\sim94\%$ model sparsity, making about $4\%\sim31\%$ improvement in graph sparsity and $3\%\sim60\%$ improvement in model sparsity compared to previous SOTA~(UGS).
At the same time, Figure~\ref{fig:figs3} shows the accuracy variation during the process of iterative magnitude-based pruning. As shown in  Figure~\ref{fig:figs3}, ACE-GLT can achieve up to 1.8 ROC-AUC improvement compared to the dense graph neural network trained on the full graph.

\subsection{Large-scale Datasets}
\begin{figure*}[!ht]
  \centering
  \includegraphics[width=1.0\linewidth, height=0.25\textheight]{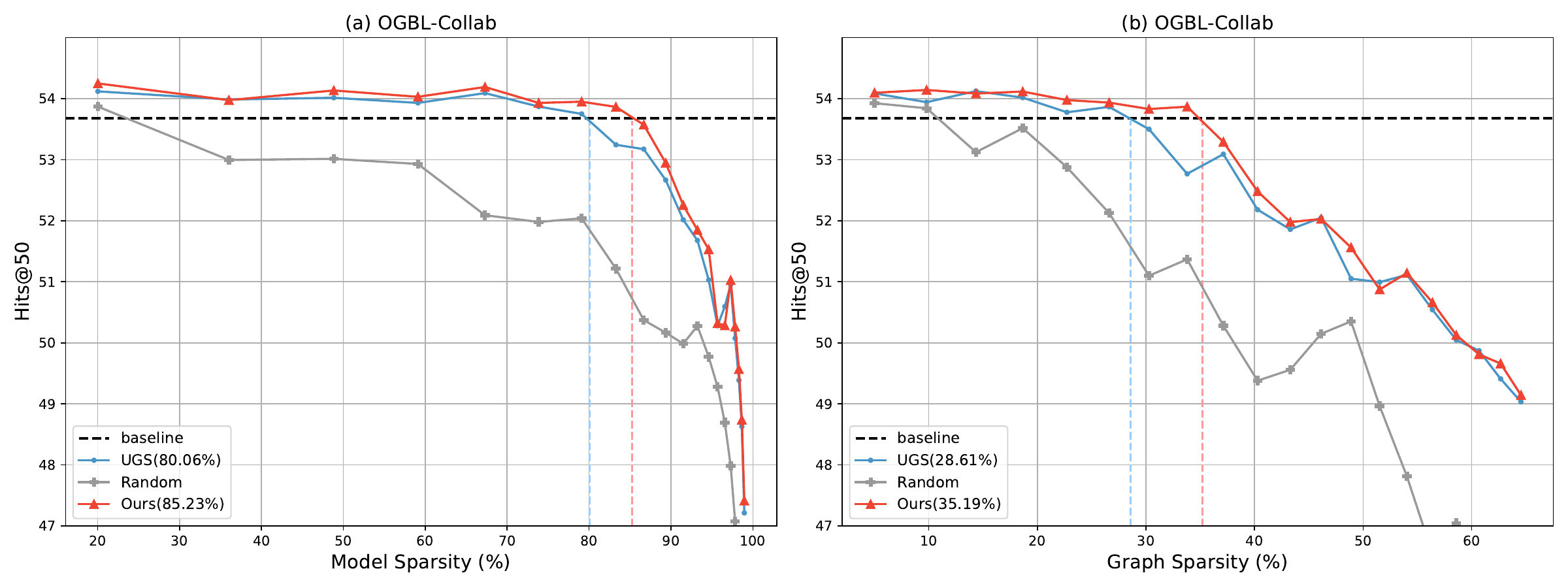}
  \captionsetup{justification=centering}
  \caption{Experimental results on the link prediction tasks of large-scale datasets. Experiments are conducted on the large-scale dataset~(ogbl-collab) with deeper GCN~(28-layer ResGCN). }
  \label{fig:figs4}
\end{figure*}

As for the node classification task, we extend our approach to the large-scale dataset and deep graph neural network on the link prediction tasks as well. The results are shown in Figure~\ref{fig:figs4}. We conduct experiments on ogbl-collab with ResGCN. And we observe that ACE-GLT can also identify sparser substructures with no performance degradation compared to UGS and find the graph lottery tickets with 85.23\% model sparsity and 35.19\% graph sparsity, which outperforms UGS by 7\% in graph sparsity and 5\% in model sparsity. These results prove that our method can significantly reduce computational costs and storage memory.

\section{Experiments under Inductive Setting}
The previously discussed experiments are based on transductive setting where all nodes are available in the training stage, however, the learned sparse substructure can also be used for unseen nodes in the inductive setting, simplifying the computational cost of aggregating feature information from their local neighborhood nodes. We construct three datasets from the Cora, Citeseer, Pubmed and OGB datasets used for inductive node classification as in~\cite{hamilton2017inductive}, then use the learned sparse substructure for testing. The baselines in non-pruned cases are shown in Table~\ref{tables5}, and the effect of GLTs found with different methods is shown in Table~\ref{tables6} and~\ref{tables7}.

\section{Additional Ablation Study}
In this section, we first add more experiments for node classification task under different fixed graph/model sparsity, and explore the effects of different fixed graph/model sparsity on the experimental effects in link prediction task.
\subsection{Different Sparsity Settings}
\begin{figure*}[h]
  \centering
  \includegraphics[width=1.0\linewidth,height=0.9\textheight]{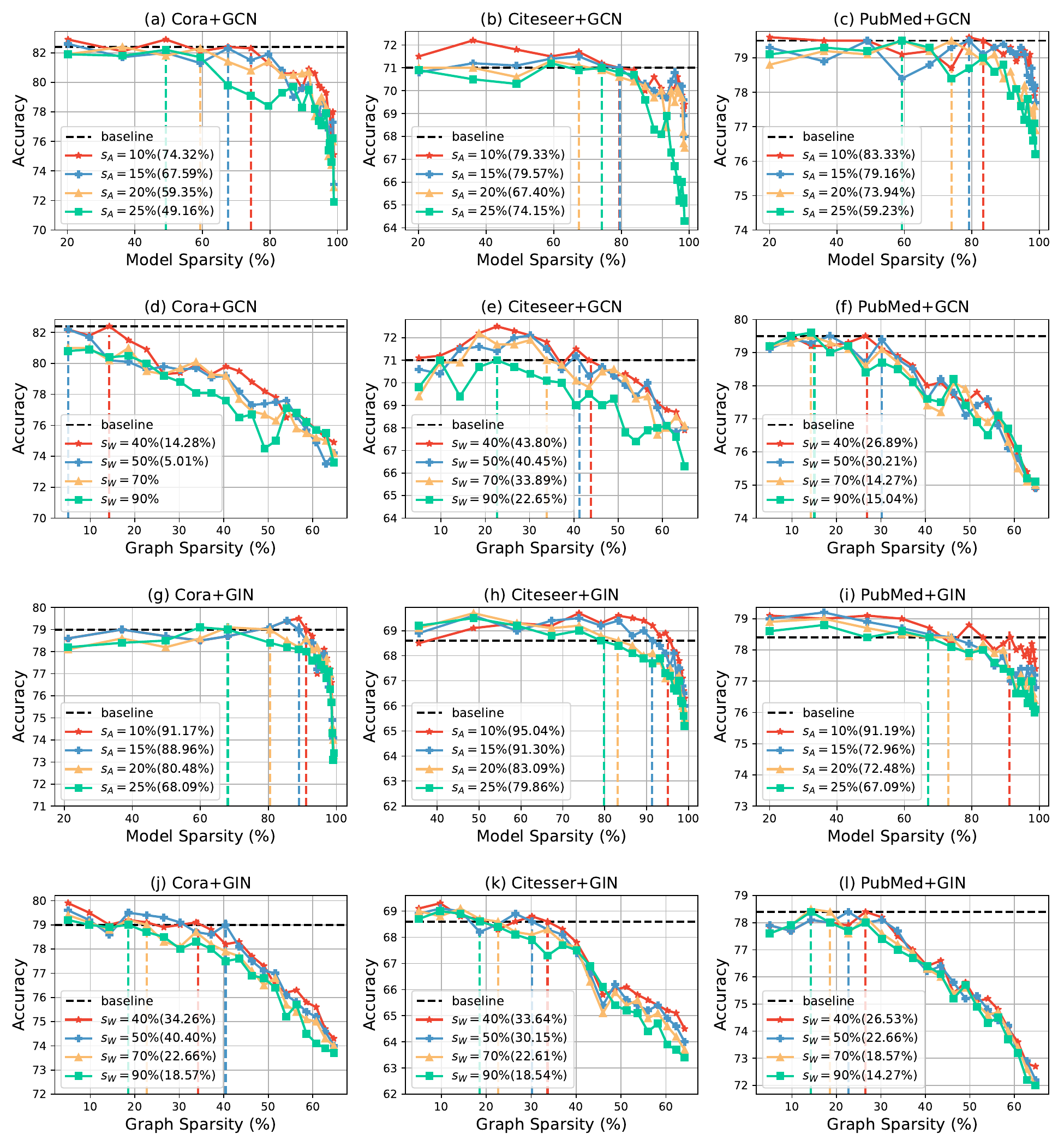}
  \captionsetup{justification=centering}
  \caption{Experimental results on link prediction tasks of small-scale datasets with different graph sparsity or model sparsity. We conduct experiments on Cora/Citeseer/PubMed with GCN/GIN. }
  \label{fig:figs5}
\end{figure*}

\begin{figure*}[h]
  \centering
  \includegraphics[width=1.0\linewidth]{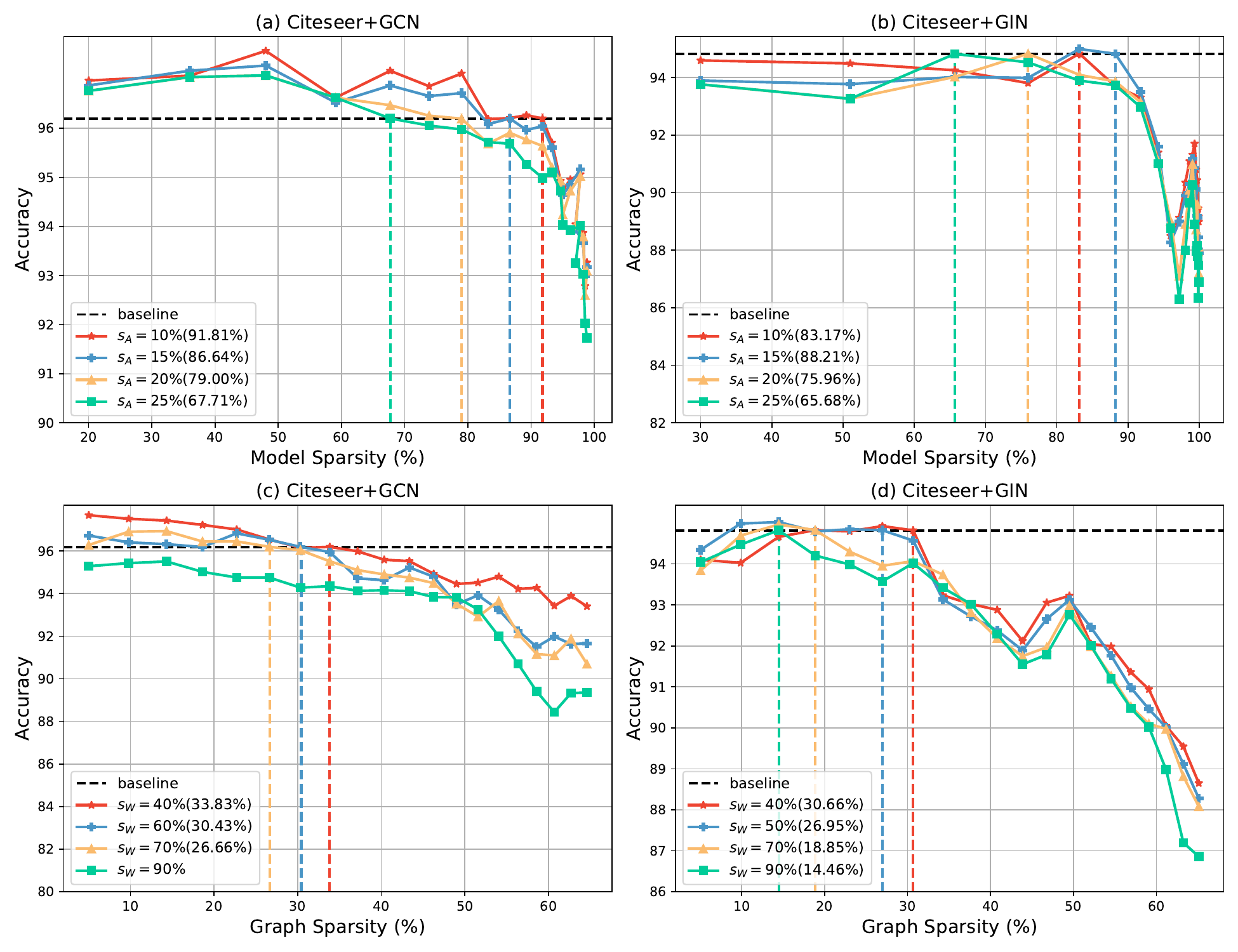}
  \captionsetup{justification=centering}
  \caption{Experimental results on link prediction tasks of small-scale datasets with different graph sparsity or model sparsity. We conduct experiments on Citeseer with GCN/GIN.}
  \label{fig:figs6}
\end{figure*}

As shown in Figure~\ref{fig:figs5}, we adjust different fixed graph sparsity~($s_A$=10\%, 15\%, 20\%, 25\%) and fixed model sparsity~($s_W$=40\%, 50\%, 70\%, 90\%) to compare the trend of accuracy when the other item is gradually increased. We provide the experiment results of GCN/GIN on Cora/Citeseer/PubMed.
Similarly, in link prediction tasks, we also apply different fixed graph sparsity~($s_A$=10\%, 15\%, 20\%, 25\%) and fixed model sparsity~($s_W$=40\%, 50\%, 70\%, 90\%) to compare the trend of accuracy when the other item is gradually increased. We provide the experiment results of GCN/GIN on Citeseer.
The results are shown in Figure~\ref{fig:figs6}.
\subsection{Re-sampling and Adaptive Sampling}
To further demonstrate the importance of re-sampling and adaptive $\mathcal{K}$, we supplement the experimental results with Cora+GIN for the node classification tasks and Cora+GCN/Citeseer+GCN for the link prediction tasks.
Similarly, we compare our method with three variants: $(1)$ with neither of them, $(2)$ with re-sample technique~(without adaptive $\mathcal{K}$), and $(3)$ with adaptive $\mathcal{K}$~(without re-sample technique).
The results are shown in Table~\ref{tables4}

\section{Time Consumption}
Our method takes longer training time (by about 2x) than others to find a better GLT. However, this extra time consumption is worthwhile since it yields substantial saving in inference time. The specifics of the training time spent by the different methods and the MACs in the inference phase of the obtained GLTs are displayed in Table~\ref{tables8}.

\begin{table*}[h]
\captionsetup{justification=centering}
\caption{Experimental results on link prediction tasks of small-scale datasets~(Cora, Citeseer and PubMed). We choose GCN, GIN and GAT as backbones. \textbf{Bold} and \underline{underline} denote the best and the second-best results, respectively. $\pm$ corresponds to one standard deviation of the average evaluation over 3 trials. ``-'' means that GLT cannot be found. }
\renewcommand{\arraystretch}{1.5}
\resizebox{\linewidth}{!}{
\begin{tabular}{@{}ccccccccccc@{}}
\toprule
\multirow{2}{*}{\textbf{Backbone}} & \multirow{2}{*}{\textbf{Method}} & \multicolumn{3}{c}{\textbf{Model Sparsity}}                                                   & \multicolumn{3}{c}{\textbf{Graph Sparsity}}                                                   & \multicolumn{3}{c}{\textbf{Highest ROC-AUC}}                                                  \\ \cmidrule(l){3-11} 
                                    &                                   & \textbf{Cora}                 & \textbf{Citeseer}             & \textbf{PubMed}               & \textbf{Cora}                 & \textbf{Citeseer}             & \textbf{PubMed}               & \textbf{Cora}                 & \textbf{Citeseer}             & \textbf{PubMed}               \\ \midrule
\multirow{3}{*}{\textbf{GCN}}       & \textbf{Random}                   & -                             & -                             & 44.56 $\pm$ 0.27 & -                             & -                             & 9.86 $\pm$ 0.14  & 91.34 $\pm$ 0.46 & 95.41 $\pm$ 0.33 & 92.57 $\pm$ 0.88 \\
                                    & \textbf{UGS}                      & \underline{71.14 $\pm$ 0.36} & -                             & \underline{83.61 $\pm$ 0.63} & \underline{37.33 $\pm$ 0.25} & 30.36 $\pm$ 0.13 & \underline{30.38 $\pm$ 0.67} & \underline{92.03 $\pm$ 0.18} & \underline{95.68 $\pm$ 0.67} & \textbf{93.39 $\pm$ 0.43} \\
                                    & \textbf{Ours}                     & \textbf{91.49 $\pm$ 0.65} & \textbf{59.38 $\pm$ 0.37} & \textbf{86.87 $\pm$ 0.21} & \textbf{53.69 $\pm$ 0.31} & \textbf{34.04 $\pm$ 0.21} & 34.14 $\pm$ 1.17 & \textbf{92.97 $\pm$ 0.12} & \textbf{96.14 $\pm$ 0.48} & \underline{93.27 $\pm$ 0.64} \\ \midrule
\multirow{3}{*}{\textbf{GIN}}       & \textbf{Random}                   & -                             & -                             & 5.29 $\pm$ 0.43  & 5.04 $\pm$ 0.32  & -                             & 26.71 $\pm$ 0.89 & 85.88 $\pm$ 0.37 & \underline{94.53 $\pm$ 0.32} & 89.62 $\pm$ 0.54 \\
                                    & \textbf{UGS}                      & \underline{85.04 $\pm$ 0.96} & \underline{78.89 $\pm$ 0.49} & \underline{74.44 $\pm$ 0.89} & \underline{25.20 $\pm$ 0.71} & \underline{26.48 $\pm$ 0.42} & \underline{30.11 $\pm$ 1.27} & \underline{86.14 $\pm$ 0.23} & 94.36 $\pm$ 0.19 & \underline{90.04 $\pm$ 0.33} \\
                                    & \textbf{Ours}                     & \textbf{89.91 $\pm$ 0.38} & \textbf{81.66 $\pm$ 0.79} & \textbf{83.15 $\pm$ 0.77} & \textbf{56.42 $\pm$ 0.48} & \textbf{36.97 $\pm$ 0.43} & \textbf{37.42 $\pm$ 0.96} & \textbf{86.89 $\pm$ 0.46} & \textbf{94.62 $\pm$ 0.96} & \textbf{90.32 $\pm$ 0.17} \\ \midrule
\multirow{3}{*}{\textbf{GAT}}       & \textbf{Random}                   & 5.19 $\pm$ 0.14  & 79.91 $\pm$ 2.36 & 5.17 $\pm$ 0.23  & 14.39 $\pm$ 0.21 & -                             & 23.31 $\pm$ 0.18 & 90.33 $\pm$ 0.16 & 94.73 $\pm$ 0.16 & 93.11 $\pm$ 0.45 \\
                                    & \textbf{UGS}                      & \underline{89.63 $\pm$ 0.27} & \underline{89.72 $\pm$ 1.38} & \underline{74.33 $\pm$ 0.68} & \underline{22.25 $\pm$ 0.79} & \underline{15.22 $\pm$ 0.71} & \underline{30.01 $\pm$ 0.45} & \underline{90.49 $\pm$ 0.57} & \underline{95.26 $\pm$ 0.24} &\underline{ 93.94 $\pm$ 0.64} \\
                                    & \textbf{Ours}                     & \textbf{93.81 $\pm$ 0.27} & \textbf{93.23 $\pm$ 0.48} & \textbf{91.76 $\pm$ 0.33} & \textbf{33.66 $\pm$ 0.31} & \textbf{30.30 $\pm$ 1.21} & \textbf{40.18 $\pm$ 0.26} & \textbf{91.04 $\pm$ 0.28} & \textbf{95.37 $\pm$ 0.11} & \textbf{94.02 $\pm$ 0.47} \\ \bottomrule
\end{tabular}}
\label{tables3}
\end{table*}

\begin{table*}[!ht]
\huge
  \captionsetup{justification=centering}
  \caption{Experimental results of ablation study on re-sample and adaptive $\mathcal{K}$ techniques. For node classification task, we conduct experiments on Cora+GIN. For link prediction task, we conduct experiments on Cora+GCN and Citeseer+GCN.}
\renewcommand{\arraystretch}{1.4}
\resizebox{\linewidth}{!}{
  \begin{tabular}{@{}cccccccccccccccc@{}}
  \toprule
  \multirow{2}{*}{}             & \multirow{2}{*}{\textbf{Task}}                & \multicolumn{2}{c}{\textbf{Setting}} & \multicolumn{6}{c}{\textbf{Model Sparsity}}            & \multicolumn{6}{c}{\textbf{Graph Sparsity}}            \\ \cmidrule(l){3-16} 
                                &                                      & \textbf{re-sampling}   & \textbf{adaptive-$\mathcal{K}$}   & \textbf{20\%}  & \textbf{40\%}  & \textbf{50\%}  & \textbf{60\%}  & \textbf{70\%}  & \textbf{80\%}  & \textbf{10\%}  & \textbf{20\%}  & \textbf{30\%}  & \textbf{40\%}  & \textbf{50\% } & \textbf{60\%}  \\ \midrule
  \multirow{4}{*}{\textbf{Cora+GIN}}     & \multirow{4}{*}{Node Classification} & $\times$      & $\times$     & 78.3  & 77.6  & 78.2  & 78.0  & 78.4  & 78.6  & 79.8  & 79.4  & 78.9  & 78.3  & 77.0  & 75.2  \\
                                &                                      & $\checkmark$  & $\times$     & 78.4  & 77.9  & 78.5  & 78.2  & 78.8  & 79.2  & 80.1  & 79.7  & 79.1  & 78.4  & 77.0  & 75.6  \\
                                &                                      & $\times$      & $\checkmark$ & 78.4  & 77.8  & 78.6  & 78.3  & 78.4  & 78.8  & 80.1  & 79.5  & 78.8  & 78.3  & 77.1  & 75.4  \\
                                &                                      & $\checkmark$  & $\checkmark$ & 78.6  & 77.6  & 78.6  & 78.2  & 78.7  & 79.7  & 80.2  & 79.6  & 79.0  & 78.5  & 77.2  & 75.8  \\ \midrule
  \multirow{4}{*}{\textbf{Cora+GCN}}     & \multirow{4}{*}{Link Prediction}   & $\times$      & $\times$     & 93.44 & 93.12 & 93.32 & 92.71 & 92.17 & 92.61 & 91.02 & 92.50 & 91.21 & 92.31 & 91.44 & 88.64    \\
                                &                                      & $\checkmark$  & $\times$     & 93.47 & 93.44 & 93.56 & 92.79 & 92.21 & 92.62 & 91.07 & 92.53 & 91.28 & 92.33 & 91.47 & 88.70 \\
                                &                                      & $\times$      & $\checkmark$ & 93.59 & 93.23 & 93.45 & 92.82 & 92.23 & 92.64 & 91.10 & 92.56 & 91.32 & 92.36 & 91.45 & 88.65 \\
                                &                                      & $\checkmark$  & $\checkmark$ & 93.62 & 93.47 & 93.67 & 92.99 & 92.24 & 92.69 & 91.09 & 92.59 & 91.35 & 92.43 & 91.49 & 88.67 \\ \midrule
  \multirow{4}{*}{\textbf{Citeseer+GCN}} & \multirow{4}{*}{Link Prediction}   & $\times$      & $\times$     & 96.61 & 96.95 & 96.21 & 96.32 & 96.72 & 96.48 & 96.92 & 96.08 & 96.11 & 96.95 & 95.50 & 94.42    \\
                                &                                      & $\checkmark$  & $\times$     & 96.63 & 96.98 & 96.24 & 96.39 & 96.71 & 96.50 & 96.91 & 96.12 & 96.15 & 96.98 & 95.51 & 94.41 \\
                                &                                      & $\times$      & $\checkmark$ & 96.64 & 96.92 & 96.27 & 96.37 & 96.74 & 96.47 & 96.91 & 96.10 & 96.14 & 97.01 & 95.53 & 94.44 \\
                                &                                      & $\checkmark$  & $\checkmark$ & 96.65 & 97.01 & 96.28 & 96.42 & 96.78 & 96.52 & 96.95 & 96.12 & 96.17 & 97.02 & 95.56 & 94.43 \\ \bottomrule
  \end{tabular}}
  \label{tables4}
\end{table*}

\begin{table*}[h]
\captionsetup{justification=centering}
\caption{Baselines under inductive setting. }
\renewcommand{\arraystretch}{1.5}
\resizebox{\linewidth}{!}{
\begin{tabular}{@{}cccccccccccc@{}}
\toprule
\multirow{2}{*}{\textbf{Metric}} & \multicolumn{3}{c}{\textbf{GCN}}                    & \multicolumn{3}{c}{\textbf{GIN}}                    & \multicolumn{3}{c}{\textbf{GAT}}                    & \multicolumn{2}{c}{\textbf{ResGCN}}          \\ \cmidrule(l){2-12} 
                                 & \textbf{Cora} & \textbf{Citeseer} & \textbf{PubMed} & \textbf{Cora} & \textbf{Citeseer} & \textbf{PubMed} & \textbf{Cora} & \textbf{Citeseer} & \textbf{PubMed} & \textbf{Ogbn-arxiv} & \textbf{Ogbn-proteins} \\ \midrule
\textbf{Accuracy}                & 75.59         & 67.98             & 74.18           & 72.12         & 68.05             & 73.29           & 78.90         & 69.65             & 73.07           & 60.62               & 77.47                  \\ \bottomrule
\end{tabular}}
\label{tables5}
\end{table*}

\begin{table*}[h]
\captionsetup{justification=centering}
\caption{Experimental results on inductive node classification tasks of small-scale datasets~(Cora, Citeseer and PubMed). We choose GCN, GIN and GAT as backbones. \textbf{Bold} and \underline{underline} denote the best and the second-best results, respectively. $\pm$ corresponds to one standard deviation of the average evaluation over 3 trials. ``-'' means that GLT cannot be found. }
\renewcommand{\arraystretch}{1.5}
\resizebox{\linewidth}{!}{
\begin{tabular}{@{}ccclccclccclcc@{}}
\toprule
\multirow{2}{*}{\textbf{Backbone}} & \multirow{2}{*}{\textbf{Method}} & \multicolumn{4}{c}{\textbf{Model Sparsity}}                                                                                                  & \multicolumn{4}{c}{\textbf{Graph Sparsity}}                                                                                                  & \multicolumn{4}{c}{\textbf{Highest Accuracy}}                                                                                                \\ \cmidrule(l){3-14} 
                                    &                                   & \multicolumn{2}{c}{\textbf{Cora}}                          & \textbf{Citeseer}                      & \textbf{PubMed}                        & \multicolumn{2}{c}{\textbf{Cora}}                          & \textbf{Citeseer}                      & \textbf{PubMed}                        & \multicolumn{2}{c}{\textbf{Cora}}                          & \textbf{Citeseer}                      & \textbf{PubMed}                        \\ \midrule
\multirow{4}{*}{\textbf{GCN}}       & \textbf{Random}                   & \multicolumn{2}{c}{-}                                      & -                                      & 72.18 $\pm$ 4.14          & \multicolumn{2}{c}{-}                                      & -                                      & 10.78 $\pm$ 0.95          & \multicolumn{2}{c}{-}                                      & -                                      & 74.32 $\pm$ 0.32          \\
                                    & \textbf{UGS}                      & \multicolumn{2}{c}{80.39 $\pm$ 3.18}          & \underline{73.31 $\pm$ 2.45}          & \underline{80.94 $\pm$ 3.83}          & \multicolumn{2}{c}{\textbf{16.71 $\pm$ 0.19}} & \underline{16.42 $\pm$ 1.26}          & 15.63 $\pm$ 1.71          & \multicolumn{2}{c}{\textbf{76.33 $\pm$ 0.08}} & \underline{68.11 $\pm$ 0.14}          & 75.08 $\pm$ 0.19          \\
                                    & \textbf{GEBT}                     & \multicolumn{2}{c}{\underline{82.98 $\pm$ 3.52}}          & 71.10 $\pm$ 3.92          & 79.86 $\pm$ 2.11          & \multicolumn{2}{c}{\textbf{12.82 $\pm$ 2.71}}          & 10.01 $\pm$ 0.01          & \underline{16.77 $\pm$ 2.82}          & \multicolumn{2}{c}{76.19 $\pm$ 0.26}          & \textbf{68.21 $\pm$ 0.24} & \underline{75.22 $\pm$ 0.24}          \\
                                    & \textbf{Ours}                     & \multicolumn{2}{c}{\textbf{90.76 $\pm$ 1.46}} & \textbf{77.12 $\pm$ 3.73} & \textbf{82.69 $\pm$ 1.93} & \multicolumn{2}{c}{12.65 $\pm$ 3.13}          & \textbf{30.28 $\pm$ 3.07} & \textbf{20.29 $\pm$ 1.43} & \multicolumn{2}{c}{\underline{76.21 $\pm$ 0.17}}          & 68.00 $\pm$ 0.06          & \textbf{75.89 $\pm$ 0.12} \\ \midrule
\multirow{4}{*}{\textbf{GIN}}       & \textbf{Random}                   & \multicolumn{2}{c}{68.09 $\pm$ 2.37}          & -                                      & -                                      & \multicolumn{2}{c}{\underline{18.87 $\pm$ 2.96}}          & -                                      & -                                      & \multicolumn{2}{c}{\textbf{73.45 $\pm$ 0.21}} & -                                      & -                                      \\
                                    & \textbf{UGS}                      & \multicolumn{2}{c}{71.24 $\pm$ 3.22}          & \underline{90.28 $\pm$ 1.98}          & 77.69 $\pm$ 5.17          & \multicolumn{2}{c}{16.34 $\pm$ 1.76}          & \underline{15.47 $\pm$ 0.19}          & \textbf{11.45 $\pm$ 0.86} & \multicolumn{2}{c}{72.87 $\pm$ 0.18}          & 68.08 $\pm$ 0.05          & 73.57 $\pm$ 0.08          \\
                                    & \textbf{GEBT}                     & \multicolumn{2}{c}{\underline{92.07 $\pm$ 1.25}}          & 63.18 $\pm$ 5.85          & \underline{80.19 $\pm$ 3.32}          & \multicolumn{2}{c}{17.31 $\pm$ 3.46}          & 12.33 $\pm$ 3.38          & \underline{6.87 $\pm$ 1.13}           & \multicolumn{2}{c}{72.85 $\pm$ 0.15}          & \underline{68.45 $\pm$ 0.11}          & \textbf{74.28 $\pm$ 0.17} \\
                                    & \textbf{Ours}                     & \multicolumn{2}{c}{\textbf{92.62 $\pm$ 1.17}} & \textbf{93.03 $\pm$ 1.10} & \textbf{79.89 $\pm$ 2.17} & \multicolumn{2}{c}{\textbf{40.05 $\pm$ 2.47}} & \textbf{22.04 $\pm$ 3.46} & 6.21 $\pm$ 0.96           & \multicolumn{2}{c}{\underline{73.24 $\pm$ 0.23}}          & \textbf{69.85 $\pm$ 0.50} & \underline{74.19 $\pm$ 0.09}          \\ \midrule
\multirow{4}{*}{\textbf{GAT}}       & \textbf{Random}                   & \multicolumn{2}{c}{40.05 $\pm$ 0.17}          & 20.25 $\pm$ 0.11          & -                                      & \multicolumn{2}{c}{12.57 $\pm$ 3.36}          & 6.08 $\pm$ 0.06           & -                                      & \multicolumn{2}{c}{78.93 $\pm$ 0.04}          & \underline{70.09 $\pm$ 0.08}          & -                                      \\
                                    & \textbf{UGS}                      & \multicolumn{2}{c}{\underline{50.82 $\pm$ 3.49}}          & 21.67 $\pm$ 0.72          & 59.08 $\pm$ 2.44          & \multicolumn{2}{c}{10.78 $\pm$ 1.04}          & 10.47 $\pm$ 0.08          & 10.11 $\pm$ 0.23          & \multicolumn{2}{c}{\textbf{79.17 $\pm$ 0.31}} & 69.76 $\pm$ 0.12          & 73.48 $\pm$ 0.21          \\
                                    & \textbf{GEBT}                     & \multicolumn{2}{c}{49.92 $\pm$ 0.73}          & \underline{36.08 $\pm$ 0.18}          & \underline{60.96 $\pm$ 1.39}          & \multicolumn{2}{c}{\underline{16.87 $\pm$ 3.02}}          & \underline{10.63 $\pm$ 1.05}          & \underline{15.87 $\pm$ 1.72}          & \multicolumn{2}{c}{78.99 $\pm$ 0.06}          & 70.04 $\pm$ 0.10          & \underline{74.03 $\pm$ 0.07}          \\
                                    & \textbf{Ours}                     & \multicolumn{2}{c}{\textbf{59.31 $\pm$ 0.74}} & \textbf{36.23 $\pm$ 0.06} & \textbf{69.57 $\pm$ 4.56} & \multicolumn{2}{c}{\textbf{17.83 $\pm$ 3.09}} & \textbf{22.50 $\pm$ 1.17} & \textbf{17.94 $\pm$ 2.32} & \multicolumn{2}{c}{\underline{79.13 $\pm$ 0.01}}          & \textbf{70.14 $\pm$ 0.06} & \textbf{74.87 $\pm$ 0.14} \\ \bottomrule
\end{tabular}}
\label{tables6}
\end{table*}

\begin{table*}[h]
\captionsetup{justification=centering}
\caption{Experimental results on inductive node classification tasks of large-scale datasets~(Ogbn-arxiv and Ogbn-proteins). We choose ResGCN as backbone. }
\resizebox{\linewidth}{!}{
\begin{tabular}{@{}cccccccc@{}}
\toprule
\multirow{2}{*}{\textbf{Backbone}} & \multirow{2}{*}{\textbf{Method}} & \multicolumn{3}{c}{\textbf{Ogbn-arxiv}}                                                       & \multicolumn{3}{c}{\textbf{Ogbn-proteins}}                                                    \\ \cmidrule(l){3-8} 
                                    &                                   & \textbf{Model Sparsity}       & \textbf{Graph Sparsity}       & \textbf{Accuracy}             & \textbf{Model Sparsity}       & \textbf{Graph Sparsity}       & \textbf{ROC-AUC}              \\ \midrule
\multirow{3}{*}{\textbf{ResGCN}}    & Random                            & -                             & -                             & -                             & 23.91 $\pm$ 0.82 & 10.14 $\pm$ 1.93 & 78.64 $\pm$ 0.36 \\
                                    & UGS                               & 47.87 $\pm$ 5.41 & \textbf{15.62 $\pm$ 1.97} & 61.19 $\pm$ 0.22 & 29.18 $\pm$ 3.34 & \textbf{11.46 $\pm$ 1.52} & \textbf{79.16 $\pm$ 0.24} \\
                                    & Ours                              & \textbf{56.39 $\pm$ 3.17} & 14.77 $\pm$ 0.91 & \textbf{62.08 $\pm$ 0.13} & \textbf{34.67 $\pm$ 2.79} & 10.93 $\pm$ 1.17 & 79.09 $\pm$ 0.13 \\ \bottomrule
\end{tabular}
}
\label{tables7}
\end{table*}

\begin{table*}[h]
\captionsetup{justification=centering}
\caption{Comparison of our method with previous SOTA~(UGS) in terms of training time and inference MACs. }
\renewcommand{\arraystretch}{1.5}
\resizebox{\linewidth}{!}{
\begin{tabular}{ccccccccc}
\hline
\multirow{2}{*}{\textbf{Backbones}} & \multirow{2}{*}{\textbf{Methods}} & \multirow{2}{*}{\textbf{Time Complexity}}                     & \multicolumn{2}{c}{\textbf{Cora}}                & \multicolumn{2}{c}{\textbf{Citeseer}}            & \multicolumn{2}{c}{\textbf{PubMed}}              \\ \cmidrule(l){4-9} 
                                    &                                   &                                                               & \textbf{Training Time} & \textbf{Inference MACs} & \textbf{Training Time} & \textbf{Inference MACs} & \textbf{Training Time} & \textbf{Inference MACs} \\ \midrule
\multirow{2}{*}{GCN}                & UGS                               & $O(k_1 \cdot L\cdot N^2 \cdot W \cdot F)$                              & 3015s                  & 863M                    & 3492s                  & 4023M                   & 19554s                 & 225M                    \\
                                    & Ours                              & $O(k_2 \cdot L \cdot N^2 \cdot W \cdot F+N^2+W)$       & 4635s                  & 208M                    & 5704s                  & 1019M                   & 23245s                 & 139M                    \\ \midrule
\multirow{2}{*}{GIN}                & UGS                               & $O(k_1 \cdot L \cdot N \cdot W \cdot F)$                                                 & 2830s                  & 201M                    & 3875s                  & 222M                    & 14061s                 & 398M                    \\
                                    & Ours                              & $O(k_2 \cdot L \cdot N \cdot W \cdot F+N^2+W)$                          & 4377s                  & 113M                    & 5897s                  & 108M                    & 23190s                 & 252M                    \\ \midrule
\multirow{2}{*}{GAT}                & UGS                               & $O(k_1 \cdot L\cdot (N\cdot F+W \cdot N \cdot F^2))$                        & 42771s                 & 1479M                   & 30021s                 & 3802M                   & 78012s                 & 1784M                   \\
                                    & Ours                              & $O(k_2 \cdot L\cdot (N \cdot F+W\cdot N\cdot F^2)+N^2+W)$ & 65653s                 & 1056M                   & 43950s                 & 2449M                   & 97690s                 & 1014M                   \\ \bottomrule
\end{tabular}}
\label{tables8}
\end{table*}

\end{document}